\documentclass[nohyperref]{article}

\usepackage{microtype}
\usepackage{graphicx}
\usepackage{subfigure}
\usepackage{booktabs} 

\usepackage{hyperref}

\usepackage[accepted]{icml2022}

\usepackage{amsmath}
\usepackage{amssymb}
\usepackage{mathtools}
\usepackage{amsthm}

\usepackage[capitalize,noabbrev]{cleveref}

\theoremstyle{plain}

\theoremstyle{definition}

\theoremstyle{remark}

\usepackage[textsize=tiny]{todonotes}

\usepackage{physics}

\usepackage{listings}
\usepackage{textcomp}
\usepackage{hyperref}

\usepackage{upquote}      
\usepackage{fancyvrb}      

\usepackage{makecell}

\icmltitlerunning{Stochastic Perturbations of Tabular Features}

\begin{document}

\twocolumn[
\icmltitle{Stochastic Perturbations of Tabular Features for Non-Deterministic Inference with Automunge}




\begin{icmlauthorlist}
\icmlauthor{Nicholas Teague}{comp}
\end{icmlauthorlist}

\icmlaffiliation{comp}{Automunge, Altamonte Springs, Florida, United States}

\icmlcorrespondingauthor{Nicholas Teague}{automunge@gmail.com}

\icmlkeywords{Machine Learning, ICML}

\vskip 0.3in
]



\printAffiliationsAndNotice{}  

\begin{abstract}
Injecting gaussian noise into training features is well known to have regularization properties. This paper considers noise injections to numeric or categoric tabular features as passed to inference, which translates inference to a non-deterministic outcome and may have relevance to fairness considerations, adversarial example protection, or other use cases benefiting from non-determinism. We offer the Automunge library for tabular preprocessing as a resource for the practice, which includes options to integrate random sampling or entropy seeding with the support of quantum circuits, representing a new way to channel quantum algorithms into classical learning.


\end{abstract}

\section{Introduction}

Stochasticity in the context of machine learning application can originate from many sources, often by design owing to corresponding regularization effects. This paper focuses on a particular kind of stochasticity associated with the injection of isotropic noise into the features of a tabular data set as passed to inference. Injecting noise into tabular features is not new, and [Appendix \ref{RelatedWork}] surveys several channels of inquiry dating back at least to the 1980's. However most of these investigations have focused on injecting noise into the features of a training data set for purposes of model characteristics. One of the premises of this paper is that the non-determinism realized from injecting noise into features passed to inference can realize benefits adjacent to a primary performance metric.

Noise injections to features translate inference from a deterministic operation to sampling from a distribution. And because the injections are applied to the input features, the sampled noise distributions are translated to the inference distribution as a function of network properties as opposed to what would result from applying noise directly to the output of inference, using the model as a source of probabilistic programming. Sort of a free lunch. Noise injections to neural network training features can be expected to improve a master performance metric and prepare the model for a corresponding noise applied to test data features. While mainstream practice for tabular learning commonly prefers gradient boosting, it has recently been shown that even simple multi-layer perceptron neural networks are capable of outperforming gradient boosting for tabular learning with a proper regularization tuning regime \cite{kadra2021welltuned}. As [Appendix \ref{Benchmarking}] will show through benchmarks, gradient boosting appears to be robust to a mild noise profile in inference, suggesting that noise could be integrated into existing pipelines of prior trained gradient boosting models [App \ref{H10}].


We briefly highlight here a few potential benefits for non-deterministic inference. We believe fairness considerations may be one example, as protected groups will as a result be exposed to a broader range of possible inference scenarios, each centered around the master model basis, and without eliminating trajectories \cite{rodriguez2021flexible}. Although there are other forms of mitigation available, the probabilistic aspects of non-deterministic inference can be applied without knowledge or identification of protected attributes. (If direction of transience is a concern, inference can also be run without noise to establish a border.) Another benefit is a robust protection against adversarial examples \cite{pmlr-v97-cohen19c}, as a targeted inference sample will not have prior knowledge of what perturbations will be seen by the model. We expect that there are many other scenarios and use cases where non-deterministic inference should be considered as the default. This paper's focus will primarily be for surveying practical considerations of how to implement.

One of our key contributions is the provision of the Automunge library \cite{anonymous_github} as a resource for stochastic perturbation application, which also serves as a platform for tabular learning that translates tidy data \cite{Wickham:14} to encoded dataframes for input to machine learning. 

\section{Related work}
\label{RelatedWork}

The integration of stochasticity into a neural network supervised learning workflow is a long established practice. Early examples included the use of noise injected to a hidden layer to derive a gradient signal \cite{NIPS1988_a0a080f4} as an alternative to backpropagation \cite{Rumelhart:1986we}, to induce a random field from an energy function \cite{6126242}, or markov chain monte carlo sampling to train a restricted boltzman machine \cite{RBMBook}. In modern practice stochasticity can be integrated into training in several forms. For example stochastic effects of dropout \cite{JMLR:v15:srivastava14a}, batch normalization \cite{pmlr-v37-ioffe15}, data augmentation \cite{perez2017effectiveness}, stochastic activation functions \cite{GhodsiStochasticReLU}, perturbed gradient updates \cite{pmlr-v97-zhou19d}, or even stochastic gradient descent itself \cite{pmlr-v119-smith20a} may introduce an implicit regularization into training, where implicit regularization refers to the notion that approximate computation can implicitly lead to statistical regularization \cite{NEURIPS2020_37740d59}. 

The injection of noise into training data features may realize a similar benefit. Training feature noise can reduce sensitivity to variations in input and improve generalization \cite{155944} with an effect equivalent to adding a regularization term to the loss function \cite{6796505}. Noise injections to training features result in smoother fitness landscape minima and reduced model complexity, with models that have lower frequencies in the Fourier domain \cite{NEURIPS2020_c16a5320} and an increased threshold for number of parameters needed to reach the overparameterization regime \cite{pmlr-v139-dhifallah21a} - which we speculate is associated with additional perturbation vectors causing an increase to the intrinsic dimension of the modeled transformation function in a manner similar to data augmentation’s impact to intrinsic dimension \cite{marcu2021datacentric}. 

The injection of noise into training features has been shown to improve model robustness, with reduced exposure to adversarial examples \cite{Zantedeschi2017EfficientDA, kannan2018adversarial}, and noise injected to inference features may further protect against adversarial examples \cite{pmlr-v97-cohen19c}. Adversarial examples can be considered a form of worst-case noise profile in comparison to the benign properties of random noise \cite{10.5555/3157096.3157279} and one can expect orders of magnitude more robustness to random noise than adversarial examples \cite{pmlr-v84-franceschi18a}. One of the premises of this work is that injected inference feature noise with a known isotropic profile can be mitigated with corresponding training feature noise profile to reduce performance degradation from the practice, where injected inference feature noise may thus be a robust protection against adversarial examples. In cases where noise with an unknown profile is expected in inference features, regularization can help to mitigate the impact \cite{WebbFunctional}. 

Label noise is a special case in comparison to feature noise, and may significantly increase the scale of training data needed to achieve comparable performance \cite{NgSlides}, although intentional label noise injections may have potential to help offset inherent label noise \cite{chen2021noise}. 

Integration of stochasticity can be tailored to architectures for autoencoders \cite{Poole2014AnalyzingNI}, reinforcement learning \cite{NEURIPS2019_e2ccf95a}, recurrent networks \cite{LimNoisyRecurrent}, convolutional networks \cite{18649f4a350c46ea966eae5bb59d3623}, and other domains. Stochastic noise is commonly applied as a channel for training data augmentation in the image modality \cite{Shorten2019ASO}. Feature noise in training may serve as a resource for masking training data properties from recovery in inference, aka differential privacy \cite{Dwork_McSherry_Nissim_Smith_2017}. We expect that elements of stochasticity may also be helpful for model perturbations in the aggregation of ensembles \cite{Dietterich00ensemblemethods}. 

Feature noise injections have mostly positive influence from the associated regularization effect, although with increasing overparameterization dropout may be expected as a more impactful stochastic regularizer \cite{pmlr-v139-dhifallah21a}. However it should be noted that there are potential negative impacts to fairness considerations. One possible consequence of the practice is that different segments of a feature set corresponding to categories in an adjacent protected feature may be impacted more than others owing to a diversity in segment distributions in comparison to a common noise profile, which may contribute to loss discrepancy between categories of that protected feature \cite{pmlr-v119-khani20a} without mitigation [Appendix \ref{Protectedattributes}]. Injecting noise directly into protected attributes may also have fairness consequences, although when there is a known noise profile there are methods available to mitigate within the loss function \cite{pmlr-v139-celis21a}. 

\section{Automunge}

Automunge \cite{anonymous_github} is an open source python library, available now for pip install, built on top of Pandas \cite{McKinney:10}, Numpy \cite{2020NumPy-Array}, SciKit-learn \cite{Pedregosa:11}, and Scipy \cite{SciPy2020}. It takes as input tabular data received in a tidy form, meaning one column per feature and one row per sample, and returns numerically encoded sets with infill to missing points, thus providing a push-button means to feed raw tabular data directly to machine learning. The extent of derivations may be minimal, such as numeric normalizations and categoric binarizations under automation, or may include more elaborate univariate transformations, including aggregated sets thereof. Generally speaking, the transformations are performed based on a fit to properties of features in a designated training set, and then that same basis may be used to consistently and efficiently prepare subsequent test data, as may be intended for use in inference or for additional training data preparation.

\begin{figure}[ht]
\vskip 0.2in
\begin{center}
\centerline{\includegraphics[width=0.95\columnwidth]{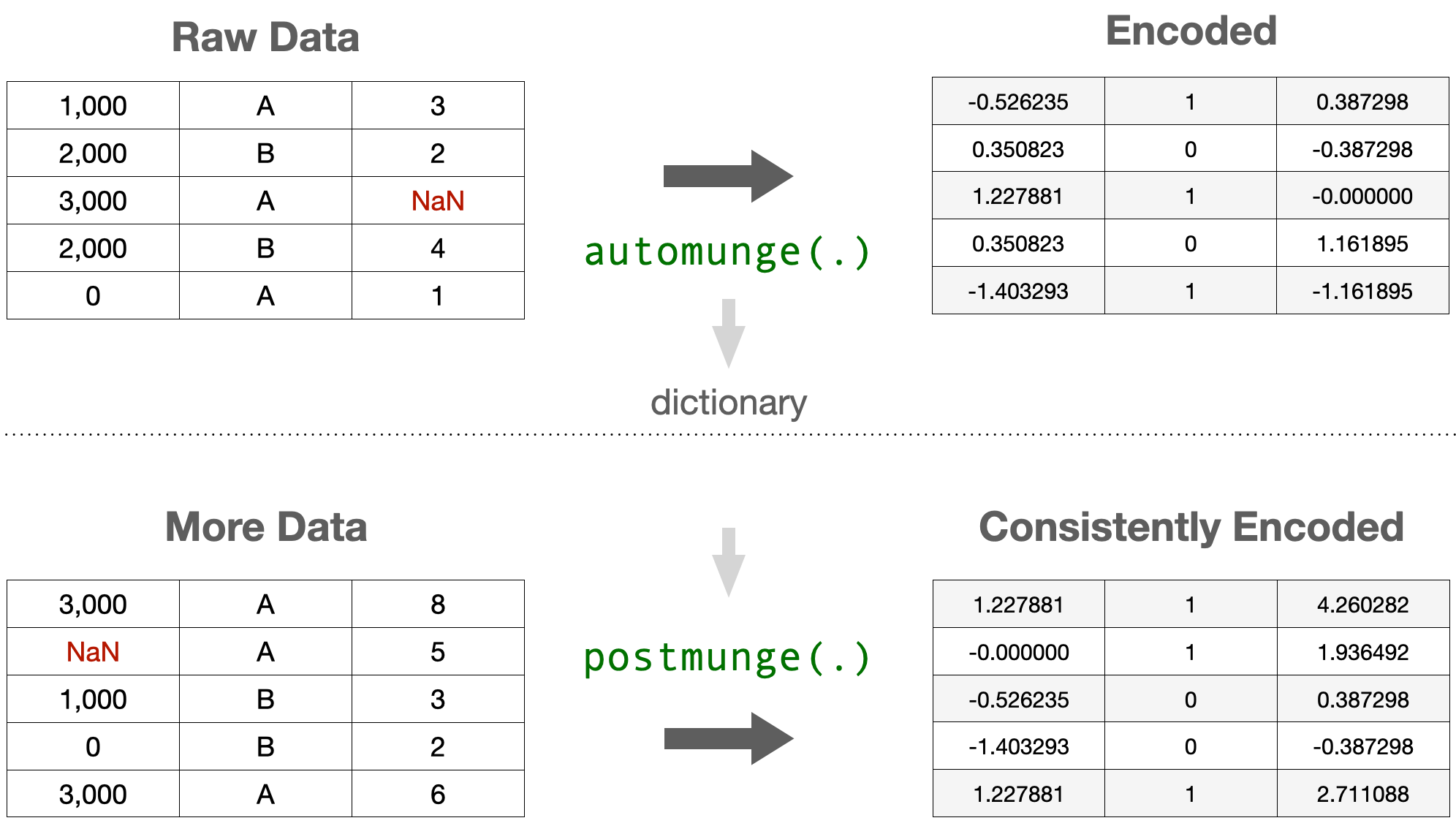}}
\caption{Automunge API process diagram}
\label{process_diagram}
\end{center}
\vskip -0.2in
\end{figure}

The interface is channeled through two master functions, automunge(.) and postmunge(.). The automunge(.) function receives a training set and if available also a consistently formatted test set, and returns a collection of dataframes intended for training, validation, and inference — each of these aggregations further segregated into subsets of features, index, and label sets. A validation set, if designated by ratio of partitioned data from the training set, is segregated from the training data prior to transformations and then consistently prepared on the train set basis to avoid data leakage between training and validation. The function also returns a populated python dictionary, which we call the postprocess\_dict, recording steps and parameters of transformations. This dictionary may then be passed along with subsequent test data to the postmunge(.) function for consistent preparations on the train set basis, as for instance may be applied sequentially to streams of data. Because it makes use of train set properties evaluated during a corresponding automunge(.) call instead of directly evaluating properties of the test data, preparing data in the postmunge(.) function can be very efficient.

There is a built in extensive library of feature encodings to choose from. Numeric features may be assigned to any range of transformations, normalizations, and bin aggregations. Sequential numeric features may be supplemented by proxies for derivatives \cite{anonymous_Numbers}. Categoric features may be encoded as ordinal, one hot, binarization, hashing, or even parsed categoric encoding \cite{anonymous_Strings} with an increased information retention in comparison to one hot encoding by a vectorization as a function of grammatical structure shared between entries. Categoric sets may be collectively aggregated into a single common binarization. Categoric labels may have label smoothing applied \cite{7780677}, or fitted smoothing where null values are fit to class distributions. Sets of transformations to be directed at targeted features can be assembled which include generations and branches of derivations by making use of our family tree primitives \cite{anonymous_github}, as can be used to redundantly encode a feature in multiple configurations of varying information content. Such transformation sets may be accessed from those predefined in an internal library for simple assignment or alternatively may be custom configured. Even the transformation functions themselves may be custom defined from a very simple template. Through application statistics of the features are recorded to facilitate detection of distribution drift. Inversion is available to recover the original form of data found preceding transformations, as may be used to recover the original form of labels after inference. Missing data is imputed by auto ML models trained on surrounding features \cite{anonymous_MissingData}.

\section{Noise Injection Options}

Noise sampling in the library is built on top of Numpy’s \cite{2020NumPy-Array} np.random module which returns an array of samples, where samples from a Bernoulli distribution could be an array of 0’s and 1’s, or samples from a Gaussian could be an array of floats. The sampling operation accepts parameters consistent with their distribution (e.g. for Gaussian mean and scale), and also a specification for shape of returned samples, which we match to the shape of the feature for targeted injection. The Appendix [Appendix \ref{A}] further surveys detail of library options, associated parameters, demonstrates application, and surveys more advanced noise profile composition techniques.

\subsection{Numeric}

For numeric injections, noise can be applied on a scaled version of the feature which allows for specification of distribution parameters independent of feature properties — e.g. when diverse numeric features are z-score normalized with a mean of 0 and a standard deviation of 1 then a noise profile can be specified independent of feature properties. In an alternate configuration, noise can be applied to an unscaled feature with noise scale adjusted based on evaluating the feature's training set distribution. For noise injection to a z-score normalized feature the multiplication of columns for sampled Bernoulli (with 0/1 entries) and Gaussian (with float entries) results in Gaussian noise injected only to entries targeted based on the Bernoulli sampling, or a common application can be applied to inject in every entry by just replacing the Bernoulli sampling with a set of 1's. The library implementation for Gaussian sampling adjacent to a Bernoulli varies slightly by only sampling entries corresponding to Bernoulli activations to reduce the entropy seeding budget.

\( (normalized) + (Bernoulli) * (Gaussian) = (injected) \)

We have not seen in prior literature the integration of a preceding Bernoulli sampling into a noise profile, so this might be considered another contribution of this paper. In benchmarking, smaller Bernoulli ratios allowed the model to tolerate a much larger scale of Gaussian noise before significant performance degradation, suggesting that in the language of information theory \cite{shannon} the entropy power of the noise distribution can be managed across both axes of scale and injection ratio.


\begin{figure}[ht]
\vskip 0.2in
\begin{center}
\centerline{\includegraphics[width=0.55\columnwidth]{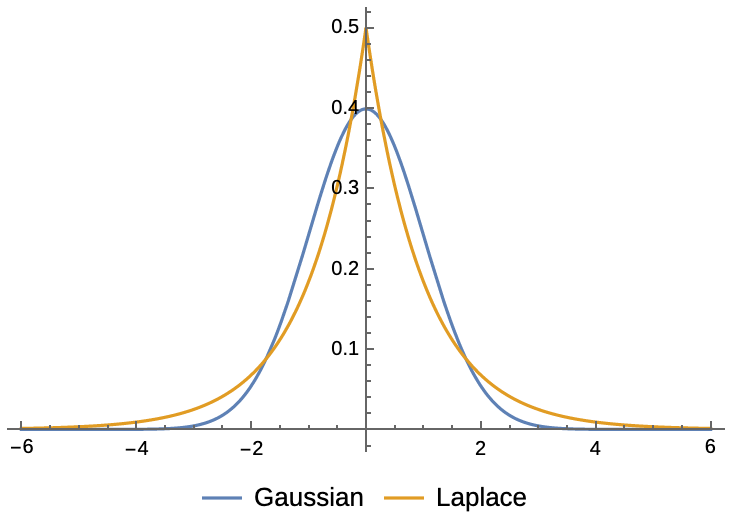}}
\caption{Probability density functions \(mean=0\), \(scale=1\)}
\label{normal_laplace}
\end{center}
\vskip -0.2in
\end{figure}

The Gaussian sampling can also be replaced with different distribution shapes, like Laplace or Uniform. The Laplace distribution, aka the double exponential distribution, has a sharper distribution peak than Gaussian, meaning more samples will be within close proximity to the mean, but also thicker tails, meaning more outlier entries may be sampled [Fig \ref{normal_laplace}]. In fact one way to think about it is that exponential tails can be considered as a kind of boundary between what would be considered thin or thick tails in other distributions \cite{cook_2019}. Thick tails refers to univariate distributions in which the aggregate measured statistics may be strongly influenced by a small number of, or even one, sampled outliers due to increased potential for extreme outliers \cite{TalebBook}, where the prototypical example would be a power law distribution. A common characteristic of thick tails are increased kurtosis which refers to steepness of peak at central mode. Injecting noise with thicker tails has benefit of exposing inference to a wider range of possible scenarios that still has noise centered to an entry, albeit at a tradeoff that number of samples needed for the noise profile to align with the target distribution climbs with tail thickness \cite{TalebHowMuchData_2019}. We expect there may prove to be benefit of other noise profile distributions in certain applications, this is an open area for future research. We did not attempt to benchmark white noise.

The numeric injections may have the potential to increase the maximum value found in the returned feature set or decrease the minimum value. In some cases, applications may benefit from retention of such feature properties before and after injection. When a feature set is scaled with a normalization that produces a known range of values, as is the case with min-max scaling (which scales data to the range between 0–1 inclusive), it becomes possible to manipulate noise as a function of entry properties to ensure retained range of values after injection [Appendix \ref{E}] [Alg \ref{alg:noisescaling}]. Other normalizations with a known range other than 0–1 (such as for `retain’ normalization \cite{anonymous_Numbers}) can be shifted to the range 0–1 prior to injection and then reverted after for comparable effect. As this results in noise distribution derived as a function of the feature distribution, the sampled noise mean can be adjusted to closer approximate a zero mean for the scaled noise [Appendix \ref{E}] [Alg \ref{alg:noisescaling_mean_adjustment}].

\subsection{Categoric}

The injection of noise into categoric features is realized by sampling from discrete distributions. For boolean integer categoric sets (as is applied to features with 2 unique values in our ‘bnry’ transform), in one configuration the injection may be applied by directly applying a Bernoulli 0/1 sample to flip targeted activations, although our base configuration applies comparable approach as ordinal encodings to take advantage of weighted replacements. 

\( abs( (boolean) - (Bernoulli) ) = (injected) \)

For categoric sets, after ordinal encoding a Bernoulli (0/1) sampling may select injection targets and a Choice sampling select alternate activations from the set of unique entries for the feature found in the training data. 

\( (ordinal) * (1 - (Bernoulli)) + (Choice) * (Bernoulli) = (injected) \)

In our base configuration, the sampling of alternate activations is weighted by frequency of those activations as were found in the training data, and number of Choice samples is instead based on number of Bernoulli activations. 

Having injected noise, a downstream transform can then be applied to convert from ordinal to some other form, or the same kind of noise can be injected downstream of a transform that reduces the number of unique entries like a hashing. We have additional configurations for categoric injections, like those applied with one hot, binarization, or multi-column hashings, that can directly be applied downstream of multi-column categoric encodings or downstream of a set of single column encodings for similar results.

\subsection{Neutral}

A type of noise profile that can be applied consistently to both numeric and categoric features is known as swap noise \cite{Ucar_SubTab}. Swap noise refers to the practice of, for Bernoulli sampled entry injection targets, replacing that entry with a Choice sampling from all entries of the same feature. One way to think about swap noise is that it is sampling from the underlying distribution of the feature. As implemented in the library, swap noise can be applied in same fashion to either single or multi column encodings. For categoric features, it has a similar result as the weighted choice sampling noted above. A tradeoff is that for injections to test data in inference the sampled swap noise is based on the test feature distribution and thus potentially exposed to covariate shift.



\section{Random Sampling}

The sampling performed by numpy.random first draws numbers from a uniform distribution which is then converted to a shaped distribution. The sampling is realized with a pseudo random number generator and currently defaults to the PCG \cite{ONeill2014PCGA} algorithm (having recently replaced the Mersenne twister \cite{10.1145/272991.272995} generator used in earlier implementations). While these types of generators are not considered on their own validated for cryptographic applications due to their use of deterministic algorithms, they have statistical properties that resemble randomness (hence the name pseudo random number generator). They generally attempt to circumvent determinism by incorporating some form of external entropy seeding into application, which in common practice is accessed from a resource in the operating system for this purpose, and may draw from such fluid channels as local memory or clock states that are known to have properties resembling randomness. In one configuration a single seed may be used to initialize a sampling operation with the remainder of samples drawn based on the progression of the generator, in an alternate configuration each sampled value may have a distinct external seed, with the second configuration having latency tradeoffs.

One of the the numpy.random generator features that the Automunge library takes advantage of is for the incorporation of supplemental entropy seeds into a sampling operation. The purpose of supplemental seeds are to access additional entropy sources other than those that may be available to the operating system for purposes of an enhanced randomness profile. Automunge has a few seeding configurations to choose from, where the quantity of seeds needed to support any of these configurations are provided to the user in a returned report [App \ref{H7}]. In one potential workflow the report may be generated by preparing a subset of the data without seeding, and then preparing the full data with seeding based on the reported quantity of seeds needed for any of these scenarios. The report details seed requirements for both train and test data so that once established the basis is known for inference. The entropy seeding scenarios are available by `sampling\_type' specification as follows:

\begin{itemize}
\item \textbf{default}: every sampling receives a common set of seeds which are shuffled and passed to each call
\item \textbf{bulk\_seeds}: every sampled entry receives a unique primary seed
\item \textbf{sampling\_seed}: every set of sampled entries receives one unique supplemental seed
\item \textbf{transform\_seed}: every noise transform receives one unique supplemental seed
\end{itemize}

In the the bulk\_seeds configuration external entropy seeds are used as the only source of seeding, as opposed to the other sampling type scenarios where it is combined as a supplemental seed to those sourced from the operating system.

As an alternative to, or in addition to, externally sampled entropy seedings, a user may also provide an alternate random number generator to be applied in place of the PCG default. The generator may be accessed from the numpy.random library \cite{2020NumPy-Array}, or may be an external library that adheres to the numpy.random.Generator format. In one configuration, the alternate generator may channel sampling operations through a quantum computing cloud resource \cite{QRAND_library}. 

In another supported configuration, a quantum computer may be accessed as an external source of entropy for supplemental seeds to be channeled through the default PCG generator, which has the benefit of intentional sampling\_type specification for budgetary considerations associated with accessing the cloud resources. When channeling external entropy seeds, the default configuration is that the seeds are added as supplemental seeds alongside those sourced from the OS, with exception of the bulk\_seeds case where external seeds are used as the only form of seeding.

We expect the contributions of Zeno gate control circuits \cite{blumenthal2021demonstration} will benefit from injecting i.i.d. stochasticity into adjacent features to increase prevalence of conical intersections, as a conservation of probability \cite{FeynmanLectures} suggests a distributional transience alignment. Injecting isotropic noise to inference allows it to be conducted in a manner benign to model performance.

\section{Conclusion}

We offer the Automunge library for tabular preprocessing as a resource for stochastic perturbations, which in addition to automating data preparations for tabular learning by numeric encodings and missing data infill also serves as a platform for engineering custom pipelines of univariate transformations fit to properties of a train set. The library includes support for channeling random sampling or entropy seeds sourced from quantum circuits into stochastic perturbations, which may benefit non-deterministic inference from a more pure randomness profile in comparison to pseudo random number generators. Full documentation and tutorials are online at \url{https://github.com/Automunge/}.

\subsubsection*{Acknowledgments}
A thank you owed to those facilitators behind Stack Overflow, Python, Numpy, Scikit-Learn, Scipy Stats, PyPI, GitHub, Colaboratory, Anaconda, VSCode, and Jupyter. A belated thank you to the communities and learning resources from TWiML and fast.ai for helping me get started. Special thanks to the team at Pandas.


\bibliography{stochastic_bib_deanon}

\bibliographystyle{icml2022}


\newpage
\appendix
\onecolumn

\section{Table of Contents}
\label{A}

\begin{itemize}

\item \textbf{Appendix B}: \hyperref[QuantumSampling]{Quantum sampling}
\item \textbf{Appendix C}: \hyperref[IntrinsicDimensions]{Intrinsic dimensions}
\item \textbf{Appendix D}: \hyperref[D]{Automunge demonstration}
\item \textbf{Appendix E}: \hyperref[FT]{Family Tree Primitives}
\item \textbf{Appendix F}:  \hyperref[B]{Train and test data}
\item \textbf{Appendix G}: \hyperref[C]{Noise options}
\item \textbf{Appendix H}: \hyperref[F]{Sampling parameters}
\item \textbf{Appendix I}: \hyperref[H]{Noise injection tutorial}
\begin{itemize}
\item \textbf{I.1}: \hyperref[H1]{DP transformation categories}
\item \textbf{I.2}: \hyperref[H2]{Parameter assignment}
\item \textbf{I.3}: \hyperref[H3]{Numeric parameters}
\item \textbf{I.4}: \hyperref[H4]{Categoric parameters}
\item \textbf{I.5}: \hyperref[H5]{Noise injection under automation}
\item \textbf{I.6}: \hyperref[H6]{Data augmentation with noise}
\item \textbf{I.7}: \hyperref[H7]{Alternate random samplers}
\item \textbf{I.8}: \hyperref[H8]{QRAND library sampling}
\item \textbf{I.9}: \hyperref[H9]{All together now}
\item \textbf{I.10}: \hyperref[H10]{Noise directed at existing data pipelines}
\end{itemize}
\item \textbf{Appendix J}: \hyperref[Ia]{Advanced noise profiles}
\begin{itemize}
\item \textbf{J.1}: \hyperref[I1]{Noise parameter randomization}
\item \textbf{J.2}: \hyperref[I2]{Noise profile composition}
\item \textbf{J.3}: \hyperref[Protectedattributes]{Protected attributes}
\end{itemize}

\item \textbf{Appendix K}: \hyperref[E]{Distribution scaling}
\item \textbf{Appendix L}: \hyperref[G]{Causal inference}
\item \textbf{Appendix M}: \hyperref[Benchmarking]{Benchmarking}
\item \textbf{Appendix N}: \hyperref[I]{Sensitivity analysis - Fastai}
\item \textbf{Appendix O}: \hyperref[J]{Sensitivity analysis - Catboost}
\item \textbf{Appendix P}: \hyperref[K]{Sensitivity analysis - XGBoost}
\item \textbf{Appendix Q}: \hyperref[L]{Intellectual property disclaimer}

\end{itemize}

\newpage

\section{Quantum sampling}
\label{QuantumSampling}

Quantum computers \cite{NielsenChuang} have an inherent advantage over classical resources for accessing more pure random states, without the need to overcome the deterministic nature of pseudo random number generation or whatever structures might be found in sources of entropy seeding. Put differently, quantum entropy is closer to i.i.d. \cite{PianiReport}. 

A quantum circuit is realized as a set of qubits, as are often initialized in the $\ket{0}$ state, that are manipulated into a joined superposition by application of quantum gates, and then a measurement operation returns a classical state from the qubits, with each probabilistically collapsing to either 0 or 1 as a function of the superposition and measurement angle. In the terminology of cloud vendors, each superposition preparation and corresponding set of measurements is considered a shot, which often serves as a pricing basis. In many quantum machine learning applications, such as for tuning the parameterized gates of a quantum neural network \cite{broughton2021tensorflow}, an algorithm may require a large number of shots to refine gate configurations from a gradient signal. When quantum circuits are applied for random number sampling, each sampled entry may only require a single or few shots to extract the desired output.

There are several types of quantum circuits that can be applied for sampling. In the simplest configuration known as the Hadamard Protocol, a single H gate is applied to each qubit, which places them in an equal superposition between $\ket{0}$ and $\ket{1}$ as the superposition ($\ket{0}$ + $\ket{1}$)/$\sqrt{2}$, resulting in an equal likelihood of sampling 0 or 1 for each measured qubit \cite{Li_QCSampling}. A slightly more elaborate approach is the Entanglement Protocol \cite{Jacak_entanglementprotocol}, in which a chain of bell states \cite{1935PhRv...47..777E} are prepared extending through the circuit depth for a similar uniform sample returned by all but the final parity qubit. Variations on these circuits may have redundancy for purposes of validation. A more resource intensive approach for sourcing randomness is known as the Sycamore Protocol, and involves preparing a circuit with randomized gates configurations at large depths, which protocol served as the basis for a milestone 2019 experiment demonstrating quantum supremacy over classical hardware \cite{Arute_quantum_Supremecy}.

There is one obstacle to complete i.i.d. randomness associated with quantum hardware to be aware of. The current generation of quantum computers available through cloud resources are still in the NISQ realm \cite{Preskill2018quantumcomputingin}, which refers to noisy intermediate scale quantum devices. In quantum computing, noise refers to channels of decoherence that may translate a qubit's pure quantum state into a mixed state \cite{2020_Witten_informationtheory}, which is undesirable for purposes of algorithmic fidelity. A prominent channel for decoherence may originate from the application of gates, and thus NISQ hardware may have some limit to circuit depth before noise starts to interfere with output, although such decoherence should not be a dominant feature in small circuit depths associated with random sampling. Depending on the style of qubit there will also be some fundamental coherence degradation rate. NISQ limitations can be overcome with error correction, and protocols have long been established to aggregate a set of physical qubits into a collective fault tolerant logical qubit, like the Shor code which aggregates 9 physical qubits into 1 logical qubit \cite{1995PhRvA..52.2493S}. The hurdle is that an architecture has to first reach a sufficient quality of qubit and fidelity of gates and measurements \cite{bharti2021noisy}, which thresholds vary by implementation.

\hyperref[A]{Table of Contents}
\newpage
\section{Intrinsic dimensions}
\label{IntrinsicDimensions}

One way to think about what is taking place with the noise injections is that the perturbation vectors result in an increase to the intrinsic dimensionality of the function \(f_d\) relating data to labels, where the intrinsic dimension of this function should be a subspace to the complexity of the environment. When estimating intrinsic dimension of a data set, some methods may attempt to isolate the data generating function from any adjacent sources of noise \cite{Granata_Estimation}, however the function  \(f_m\) modeled by the network will include both channels. When a deep network learns a representation of the transformation function, the model's intrinsic dimensionality can be expected to adhere closer to the dimensionality of the function as opposed to the complexity of the domain \cite{cloninger2021deep}. 

An intuitive interpretation could be that minimizing the distribution shift between \(f_d\) and \(f_m\) with a comparable intrinsic dimension may benefit performance, however the benefits of data augmentation suggest that distribution manipulations that increase the intrinsic dimension may be constructive to a model's ability to generalize \cite{marcu2021datacentric}. One of the benefits of the tabular modality in comparison to image domain is that training data sets are often capable at reasonable scales to fully capture characteristics of the function \(f_d\) available within the observed features. In a benchmarking experiment for tabular data augmentation by gaussian noise injections to training features in a deep learning application \cite{anonymous_Numbers}, it was found that such noise augmentation was mostly benign to model performance in a fully represented data set, but was increasingly beneficial as the training data scale before augmentation was reduced to simulate underrepresentation. We speculate this could be from in cases of underrepresented training data, a modeled function \(f_m\) will capture representations of spurious dimensions associated with gaps in the fidelity of \(f_d\) from underrepresented training data, and that the introduction of noise augmentation serves to dampen the modeling of spurious dimensions by diverting complexity capacity of the network from less tractable spurious dimensions associated with gaps in fidelity from underrepresented data to tractable noise dimensions. Thus even though the noise augmentation is increasing the intrinsic dimension of \(f_d\), it is doing so in a manner that improves generalization performance. Further, the types of noise applied, when in some manner aligned or correlated with the data generating function, will not have a fully additive increase to intrinsic dimension - this is why augmentation with common noise profiles like Gaussian, which arise naturally under the central limit theorem, will likely perform better than some exotic variant - similar to how in image modality is is common to augment by mirroring images between left and right as opposed to up and down.

Having established that noise injection will increase the intrinsic dimension of \(f_d\), it is interesting that even though the modeled \(f_m\) should also be higher, the resulting stochastic regularization may actually dampen dimensionality. \cite{NEURIPS2020_c16a5320} demonstrated that the effect of stochastic regularization realized from Gaussian noise can be related to the Fourier representation of the modeled function \(f_m\) by considering it's form in Sobolev space and modeling the data as a density function, which demonstrated that the noise injections to training data produces a regularization that penalizes higher frequency terms, resulting in a modeled function \(f_m\) with lower-frequency components, and further that this dampening appeared to be recursive through layers, such that deeper layers realized lower frequency components compared to earlier. Such dampening of Fourier frequencies would be consistent with a reduced intrinsic dimensionality.

The question of Fourier domain representation of the modeled function is actually relevant to other aspects of quantum computing. In variational quantum algorithms \cite{2021Nature}, parameterized gates are tuned from a gradient signal with a classical optimizer, and such parameterized quantum circuits can even be integrated as modules mixed with classical neural networks as a combined quantum neural network to be collectively trained with a common gradient signal \cite{broughton2021tensorflow}. There is an interesting distinction to be made between classical learning and variational quantum circuits associated with neutrality to data representations. In deep learning \cite{goodfellow2016deep} some people consider the practice of feature engineering as obsolete, as deep networks are universal function approximators. Performance of variational quantum circuits are more tied to data encoding strategies. The tuning of gates are actually learning Fourier coefficients, but they are limited to learning coefficients of frequencies that are represented in the data encoding space. Thus one important lever to improve the performance of variational quantum circuits is by enhancing the frequency spectrum of the data encodings \cite{Schuld2021}.


The relevance to application of stochastic perturbations is that since stochastic regularization is actually dampening the Fourier frequencies, an added benefit of non-deterministic inference is that the noise injections to test data should have the result of recovering some of the dampened frequencies in space of inference, which may become beneficial in classical machine learning applications adjacent to quantum computing applications, like for those mixed modules between classical and quantum learning that may be configured in a quantum neural network.

\hyperref[A]{Table of Contents}
\newpage
\section{Automunge demonstration}
\label{D}

The Automunge interface is channeled through two master functions, automunge(.) for preparing data and postmunge(.) for preparing additional corresponding data. As an example, for a training set dataframe df\_train which includes a label feature `labels', automunge(.) can be applied under automation as:
\begin{lstlisting}
#!pip install Automunge
from Automunge import *
am = AutoMunge()

train, train_ID, labels, \
val, val_ID, val_labels, \
test, test_ID, test_labels, \
postprocess_dict = \
am.automunge(df_train,
                 labels_column = 'labels')
\end{lstlisting}

Some of the returned sets may be empty based on parameter selections. Using the returned dictionary postprocess\_dict, corresponding data can then be prepared on a consistent basis with postmunge(.).
\begin{lstlisting}
test, test_ID, test_labels, \
postreports_dict = \
am.postmunge(postprocess_dict, 
                 df_test)
\end{lstlisting}

To engineer a custom set of transformations, one can populate a transformdict and processdict entry for a new transformation category we'll call `newt'. The functionpointer is used to match `newt' to the processdict entries applied for `nmbr', which is for z-score normalization. The transformdict is used to populate transformation category entries to the family tree primitives [Table \ref{family_tree_primitives}], [Appendix \ref{FT}] \cite{anonymous_github} associated with a root category. The first four primitives are for upstream transforms. Since parents is a primitive with offspring, after applying transforms for the `newt' entry, the downstream primitives from newt's family tree will be inspected to apply `bsor' for ordinal encoded standard deviation bins to the output of the upstream transform. The upstream `NArw' is used to aggregate missing data markers. The assigncat parameter is used to assign `newt' as a root category to a target input column `targetcolumn'. There are also many preconfigured trees available in the library.

\begin{lstlisting}
processdict =  {'newt' : {'functionpointer'   : 'nmbr'}}    

transformdict =  {'newt' : {'parents'         : ['newt'],
                                    'siblings'             : [],
                                    'auntsuncles'      : [],
                                    'cousins'             : ['NArw'],
                                    'children'             : [],
                                    'niecesnephews' : [],
                                    'coworkers'         : [],
                                    'friends'              : ['bsor']}}
                        
assigncat = {'newt' : ['targetcolumn']}
\end{lstlisting}

\begin{table}[h]
\caption{Family Tree Primitives}
\label{sample-table}
\begin{center}
\begin{tabular}{lllll}
\multicolumn{1}{c}{\bf Primitive}  &\multicolumn{1}{c}{ \thead{\bf Upstream /\\\bf Downstream}} &\multicolumn{1}{c}{\thead{\bf Applied to \\ \bf Generation}} &\multicolumn{1}{c}{\bf Column Action} &\multicolumn{1}{c}{\thead{\bf Downstream \\\bf Offspring} }
\\ \hline \\
parents & upstream & first &replace &yes\\
siblings & upstream & first &supplement &yes\\
auntsuncles  & upstream & first &replace &no\\
cousins & upstream & first &supplement &no\\
children & downstream parents & offspring &replace &yes\\
niecesnephews & downstream siblings & offspring &supplement &yes\\
coworkers & downstream auntsuncles & offspring &replace &no\\
friends & downstream cousins & offspring &supplement &no\\
\label{family_tree_primitives}
\end{tabular}
\end{center}
\end{table}

This transformation set will return columns with headers logging the applied transformation categories as: `column\_newt' (z-score normalization), `column\_newt\_bsor' (ordinal encoded standard deviation bins), and `column\_NArw' (missing data markers). In an alternate configuration `bsor' could be entered to an upstream primitive, this is just an example to demonstrate applying generations of transformations. Since friends is a supplement primitive, the upstream output `column\_newt' to which the `bsor' transform is applied is retained in the returned data. And since cousins and friends are primitives without offspring, no further generations are inspected after applying their entries.

Parameters can be passed to the transformations through assignparam, as demonstrated here for updating a parameter setting so that the number of standard deviation bins for `bsor' as applied to column `column' is increased from the default of 6 to 7, where since this is an odd number will result in the center bin straddling the mean.
\begin{lstlisting}
assignparam = {'bsor' : {'column' : {'bincount' : 7}}}
\end{lstlisting}

Under automation auto ML models are trained for each feature and missing marker activations are aggregated in the returned sets to support missing data imputation. These options can be deactivated with the MLinfill and NArw\_marker parameters. The function automatically shuffles the rows of training data and defaults to not shuffling rows of test data. To retain order of train set rows can deactivate the shuffletrain parameter.
\begin{lstlisting}
shuffletrain = False
\end{lstlisting}

There is an option to mask the returned feature headers and order of columns for purposes of retaining privacy of model basis by the privacy\_encode parameter, which can later be inverted with a postmunge(.) inversion operation if desired. This option can also be combined with encryption of the postprocess\_dict by the encrypt\_key parameter. Here is an example of privacy encoding without encryption.
\begin{lstlisting}
privacy_encode = True
\end{lstlisting}

Putting it all together in an automunge(.) call simply means passing our parameter specifications.
\begin{lstlisting}
train, train_ID, labels, \
val, val_ID, val_labels, \
test, test_ID, test_labels, \
postprocess_dict = \
am.automunge(df_train,
                 labels_column = 'labels',
                 processdict = processdict,
                 transformdict = transformdict,
                 assigncat = assigncat,
                 assignparam = assignparam,
                 shuffletrain = shuffletrain,
                 privacy_encode = privacy_encode)
\end{lstlisting}

One can then save the returned postprocess\_dict, such as by downloading with the pickle library, to use as a key for preparing additional corresponding data on a consistent basis with postmunge(.).
\begin{lstlisting}
test, test_ID, test_labels, \
postreports_dict = \
am.postmunge(postprocess_dict, 
                 df_test)
\end{lstlisting}

Assigning noise injection root categories to targeted input columns is also applied in the assigncat automunge(.) parameter, which once assigned will be carried through as the basis for postmunge(.). Here we demonstrate assigning DPnb as the root category for a list of numeric features, DPod for a list of categoric features, and DPmm for a specific targeted numeric feature.

\begin{lstlisting}
assigncat = \
{'DPnb' : numeric_features_list,
 'DPod' : categoric_features_list,
 'DPmm' : '<targetcolumn>'}
\end{lstlisting}

To default to applying noise injection under automation one can take advantage of the automunge(.) powertransform parameter which is used to select between scenarios for default transformations applied under automation. powertransform accepts specification as `DP1’ or `DP2’ resulting in automated encodings applying noise injection, further detailed in the read me powertransform parameter writeup (or DT and DB equivalents DT1 / DT2 / DB1 / DB2 for different default train and test noise configurations [Appendix \ref{B}]).

Transformation category specific parameters can be passed to transformation functions through the automunge(.) assignparam parameter, which will then be carried through as the basis for preparing additional data in postmunge. In order of precedence, parameter assignments may be designated targeting a transformation category as applied to a specific column header with suffix appenders, a transformation category as applied to an input column header (which may include multiple instances), all instances of a specific transformation category, all transformation categories, or may be initialized as default parameters when defining a transformation category.

Here we demonstrate passing three different kinds of assignparam specifications.

\begin{lstlisting}
assignparam = \
{'global_assignparam'  : 
    {'testnoise': True},
 'default_assignparam' : 
    {'DPod' : {'flip_prob' : 0.05}},
 'DPmm' : 
    {'targetcolumn' : 
        {'noisedistribution' : 'abs_normal',
         'sigma' : 0.02}}}
\end{lstlisting}

\begin{itemize}
\item \textbf{`global\_assignparam’} passes a specified parameter to all transformation functions applied to all columns, which if a function does not accept that parameter will just be ignored. In this demonstration we turn on test noise injection for all transforms via the `testnoise’ parameter.
\item \textbf{`default\_assignparam’} passes a specified parameter to all instances of a specified tree category (where tree category refers to the entries to the family tree primitives of a root category assigned to a column, and in many cases the tree category will be the same as the root category). Here we demonstrate updating the `flip\_prob’ parameter from the 0.03 default for all instances of the DPod transform, which represents the ratio of entries that will be targeted for injection.
\item To target parameters to specific categories as applied to specific columns, can specify as \textbf{\{category : \{column : \{parameter : value\}\}\}}. Here we demonstrate targeting the application of the DPmm transform to a column `targetcolumn’ in order to apply all positive signed noise injections by setting the `noisedistribution’ parameter to `abs\_normal’, and also reducing the standard deviation of the injections from default of 0.03 to 0.02 with the `sigma’ setting. `targetcolumn’ refers to the header configuration received as input to a transform without the returned suffix.
\end{itemize}

Having defined our assignparam specification dictionary, it can then be passed to the automunge(.) assignparam parameter. As an asterisk, it’s important to keep in mind that targeting a category for assignparam specification is based on that category’s use as a tree category (as opposed to use as a root category), which in some cases may be different. The read me documentation on noise injection details any cases where a noise injection parameter acceptance may be a tree category differing from the root category, as is the case for a few of the hashing noise injections. Having defined our relevant parameters, we can then pass them to an automunge(.) call.

\begin{lstlisting}

#!pip install Automunge
from Automunge import *
am = AutoMunge()

#import train and test data sets
import pandas as pd
df_train = pd.read_csv('train.csv')
df_test = pd.read_csv('test.csv')
labels_column = '<labels_column_header>'
trainID_column = '<ID_column_header>'

#prepare the data for machine learning
train, train_ID, labels, \
val, val_ID, val_labels, \
test, test_ID, test_labels, \
postprocess_dict = \
am.automunge(df_train,
                 df_test = df_test,
                 labels_column = labels_column, 
                 trainID_column = trainID_column,
                 assigncat = assigncat,
                 assignparam = assignparam)

#download postprocess_dict with pickle
\end{lstlisting}

In addition to preparing our training data and any validation or test data, this function also populates the postprocess\_dict dictionary, which we recommend downloading with pickle if you intend to train a model with the returned data (pickle code demonstrations provided in read me). The postprocess\_dict can then be uploaded in a separate notebook to prepare additional corresponding test data on a consistent basis, as may be used for inference.

\begin{lstlisting}
#!pip install Automunge
from Automunge import *
am = AutoMunge()

#import test data
import pandas as pd
df_test = pd.read_csv('test.csv')

#upload postprocess_dict with pickle
#now prepare the test data on a consistent basis
#traindata parameter accepts boolean, defaulting to False
test, test_ID, test_labels, \
postreports_dict = \
am.postmunge(postprocess_dict, 
                 df_test,
                 traindata = False)
\end{lstlisting}

\hyperref[A]{Table of Contents}

\newpage
\section{Family Tree Primitives}
\label{FT}

\begin{figure}[ht]
\vskip 0.2in
\begin{center}
\centerline{\includegraphics[width=0.95\columnwidth]{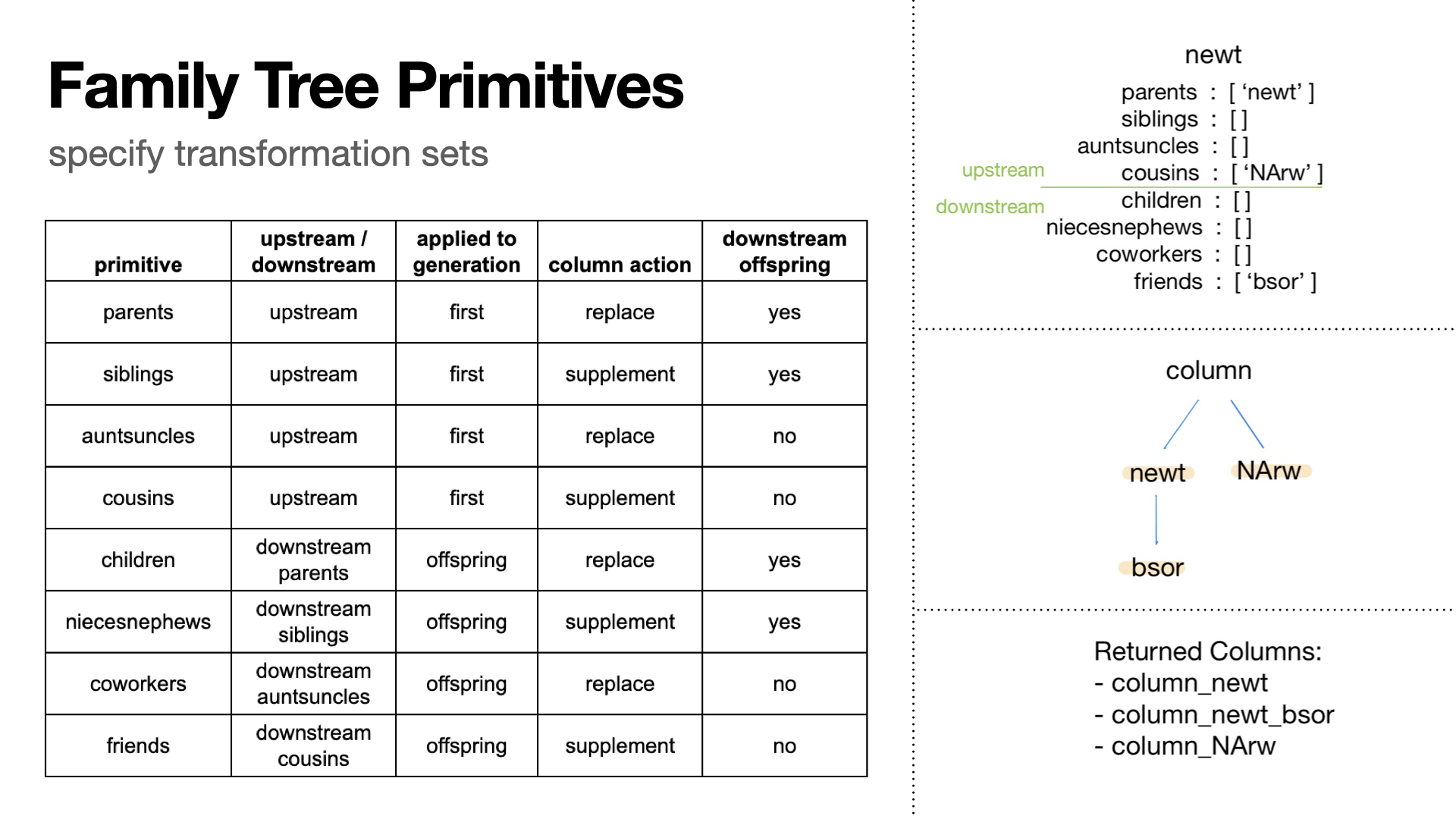}}
\caption{Automunge family tree primitives}
\label{Family_Tree_Primitives}
\end{center}
\vskip -0.2in
\end{figure}

\hyperref[A]{Table of Contents}

\newpage
\section{Train and test data}
\label{B}

One of the key distinctions in the library with respect to data preparation is the difference between training data and test data. In a traditional supervised learning application, training data would be used to train a model, and test data would be used for validation or inference. When Automunge prepares training data, in many cases it fits transformations to properties of a feature as found in the training data, which is then used for preparing that feature in the test data on a consistent basis. The automunge(.) function for initial preparations accepts training data and optionally additionally test data with application. The postmunge(.) function for preparing additional data assumes that received data is considered test data by the default parameter setting traindata=False.

In the context of noise injections, the train/test distinction comes into play. Our original default configuration was that noise is injected to training data and not injected to test data. This was built around use cases of applying noise for applications in model training, such as for data augmentation, differential privacy, and model perturbation in the aggregation of ensembles. As we'll demonstrate through benchmarking [Appendix \ref{I}, \ref{J}, \ref{K}] there are scenarios in application of non-deterministic inference for noise injection for test data as well as train data (e.g. neural networks) — or perhaps even just for test data and not for train data (e.g. gradient boosting).

We thus have a few relevant parameters for distinguishing between these scenarios. In the base configuration for `DP' root categories, the training data set returned from automunge(.) receives noise when relevant transforms are applied, and does not receive noise to the corresponding features in test data, including test data sets returned from both automunge(.) or postmunge(.). 

To treat test data passed to postmunge(.) as training data, postmunge has the traindata parameter, which can be turned on and off as desired with each postmunge call. To configure a transformation to default to applying injected noise to train or test data, parameters can be passed to specific transformations as applied to specific columns with the automunge(.) assignparam parameter. The noise injections transforms accept a trainnoise specification (defaulting as True) signaling that noise will be injected to training data, and a testnoise specification (defaulting as False) signaling that noise will not be injected to test data. Please note that these assignparam parameters, once specified in an automunge(.) call, are retained as basis for preparing additional data in postmunge(.). If validation data is prepared in automunge(.) it is treated comparably to test data. Alternate root categories are available by replacing a category's `DP' prefix with `DT' or `DB' with other train and test injection defaults as shown [Table \ref{traindata_scenarios}].

\begin{table}[h]
  \caption{Train and test injection scenarios}
  \centering
  \includegraphics[width=0.6\linewidth]{"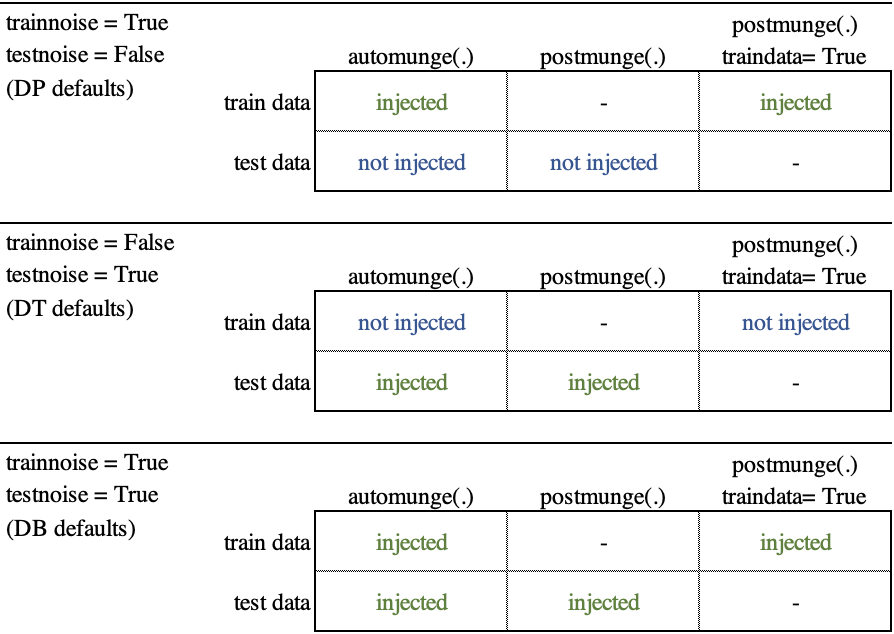}
  \label{traindata_scenarios}
\end{table}

Please note that the postmunge(.) traindata parameter can also be passed as `train\_no\_noise' or `test\_no\_noise', which is for purpose of treating the data consistent to train or test data but without noise injections.

\hyperref[A]{Table of Contents}


\newpage
\section{Noise options}
\label{C}

Noise injections in the library can be applied in conjunction with other preparations, for example noise injection for numeric features can be applied in conjunction with normalizations and scaling, or noise injections to categoric features can be applied in conjunction with integer encodings. Thus root categories for noise injection transformations [Table \ref{root_cat_table}] can be considered a drop in replacement for the corresponding encoding. As used in [Table \ref{root_cat_table}], distribution sampling refers to Gaussian noise that can be configured by parameter to alternate profiles like Laplace, Uniform, or all positive/negative.

The library also supports the application of noise injections without otherwise editing a feature. DPse (passthrough swap noise), DPpc (passthrough weighted activation flips for categoric), DPne (passthrough gaussian or laplace noise for numeric), DPsk (passthrough mask noise for numeric or categoric), and excl (passthrough without noise) can be used in tandem to pass a dataframe to automunge(.) for noise injection without other edits or infill, such as could be used to incorporate noise into an existing tabular pipeline. When limited to these three root categories the returned dataframe will match the same order of columns with only edits other than noise as updated column headers and DPne will override any data types other than float. (To retain same order of rows can deactivate shuffletrain parameter, and original column headers can be retained with the orig\_headers parameter.) This practice will be demonstrated in [Appendix \ref{H10}].



\begin{table}[!ht]
    \centering
    \caption{Noise root categories}
    \begin{tabular}{|l|l|l|l|l|}
    \hline
        \textbf{Root} & \textbf{Feature} & \textbf{Resembles} &   \textbf{Encoding} &   \textbf{Noise Type}  \\ \hline
        DPnb & numeric & nmbr &   z-score normalization &   distribution sampling  \\ \hline
        DPmm & numeric & mnmx &   min-max scaling &   scaled distribution sampling  \\ \hline
        DPrt & numeric & retn &   retain normalization &   scaled distribution sampling  \\ \hline
        DPbn & categoric & bnry &   single column boolean integer encoding &   weighted activation flip  \\ \hline
        Dpod & categoric & ord3 &   ordinal encoding &   weighted activation flip \\ \hline
        DP10 & categoric & 1010 &   binarization &   weighted activation set flip \\ \hline
        Dpoh & categoric & onht &   one-hot encoding &   weighted activation set flip \\ \hline
        DPhs & categoric & hash &   multicolumn hashing &   weighted activation flip \\ \hline
        DPh2 & categoric & hsh2 &   single column hashing &   weighted activation flip \\ \hline
        DPh1 & categoric & hs10 &   binarized multicolumn hashing &   weighted activation set flip \\ \hline
        DPne & numeric & excl &   pass-through &   distribution sampling  \\ \hline
        DPpc & categoric & excl &   pass-through &   weighted activation flip  \\ \hline
        DPse & neutral & excl &   pass-through &   swap noise  \\ \hline
        DPsk & neutral & excl &   pass-through &   mask noise \\ \hline
    \end{tabular}
    \label{root_cat_table}
\end{table}

\hyperref[A]{Table of Contents}

\newpage
\section{Sampling parameters}
\label{F}

\begin{figure}[ht]
\vskip 0.2in
\begin{center}
\centerline{\includegraphics[width=\columnwidth]{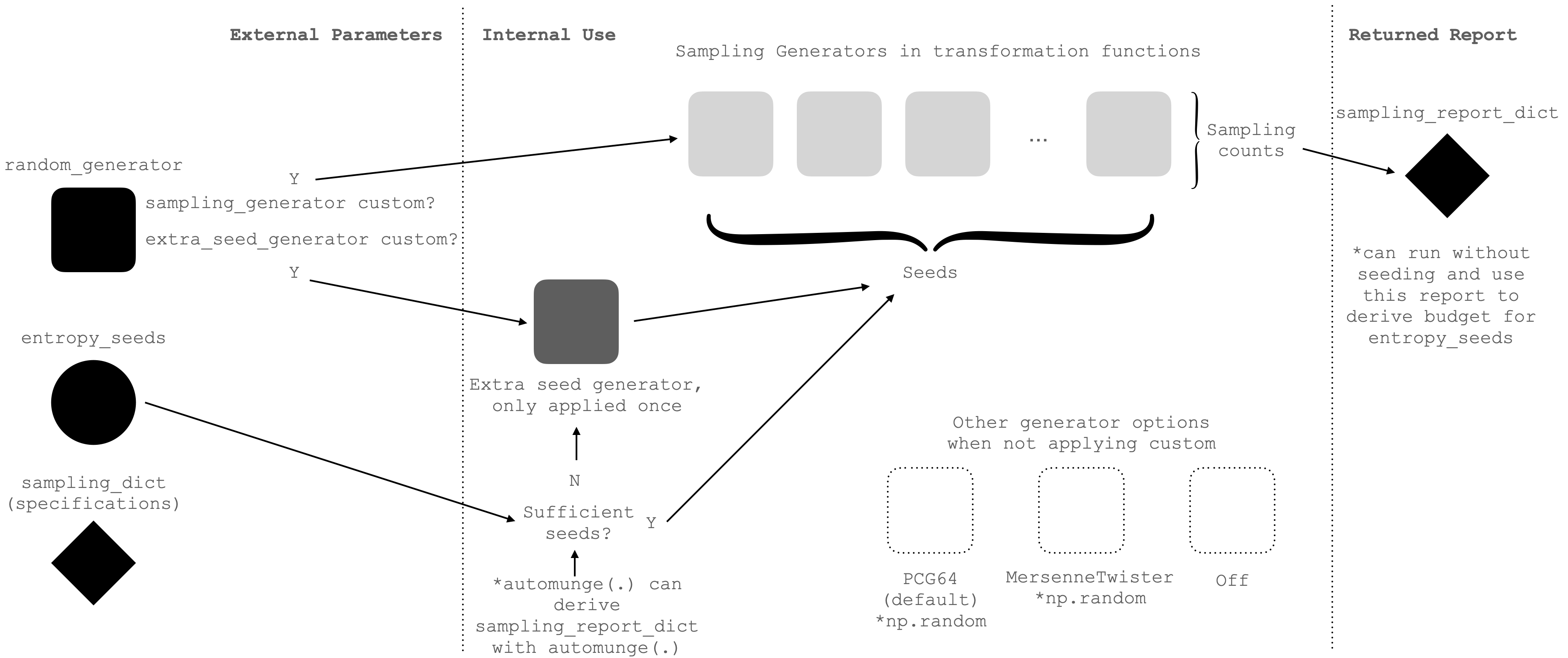}}

\caption{automunge(.) or postmunge(.) sampling parameters}
\label{sampling_parameters_figure}
\end{center}
\vskip -0.2in
\end{figure}

For entropy seeding and alternate random samplers, each automunge(.) or postmunge(.) call may be passed distinct parameters [Fig \ref{sampling_parameters_figure}], meaning that the sampling parameters passed to automunge(.) are not carried through as a basis to postmunge(.). This is to support potential workflow where entropy seeding may be desired for training and not inference or vice versa. The three relevant parameters are entropy\_seeds, random\_generator, and sampling\_dict (which aggregates multiple sub-specifications into a dictionary).

* \textbf{entropy\_seeds}: defaults to False, accepts integer or list / flattened array of integers which may serve as supplemental sources of entropy for noise injections with DP transforms, we suggest integers in range \{0:(2 ** 31 - 1)\} to align with int32 dtype. entropy\_seeds are specific to an automunge(.) or postmunge(.) call, in other words they are not returned in the populated postprocess\_dict. Please note that for determination of how many entropy seeds are needed for various sampling\_dict[`sampling\_type'] scenarios, can inspect postprocess\_dict[`sampling\_report\_dict'], where if insufficient seeds are available for these scenarios additional seeds will be derived with the extra\_seed\_generator.  Note that the sampling\_report\_dict will report requirements separately for train and test data and in the bulk\_seeds case will have a row count basis. (If not passing test data to automunge(.) the test budget can be omitted.) Note that the entropy seed budget only accounts for preparing one set of data, for the noise\_augment option we recommend passing a custom extra\_seed\_generator with a sampling\_type specification, which will result in internal samplings of additional entropy seeds for each additional noise\_augment duplicate (or for the bulk\_seeds case with external sampling can increased entropy\_seed budget proportional to the number of additional duplicates with noise).

* \textbf{random\_generator}: defaults to False, accepts numpy.random.Generator formatted random samplers which are applied for noise injections with DP transforms. Note that random\_generator may optionally be applied in conjunction with entropy\_seeds. When not specified applies numpy.random.PCG64. Examples of alternate generators could be a generator initialized with the QRAND library \cite{QRAND_library} to sample from a quantum circuit. Or if the alternate library does not have numpy.random support, their output can be channeled as entropy\_seeds for a similar benefit. random\_generator is specific to an automunge(.) or postmunge(.) call, in other words it is not returned in the populated postprocess\_dict. Please note that numpy formatted generators of both forms e.g. np.random.PCG64 or np.random.PCG64() may be passed, in the latter case any entropy seeding to this generator will be turned off automatically.

* \textbf{sampling\_dict}: defaults to False, accepts a dictionary including possible keys of \{sampling\_type, seeding\_type, sampling\_report\_dict, stochastic\_count\_safety\_factor, extra\_seed\_generator, sampling\_generator\}. sampling\_dict is specific to an automunge(.) or postmunge(.) call, in other words they are not returned in the populated postprocess\_dict. 

  - sampling\_dict[`\textbf{sampling\_type}'] accepts a string as one of \{`default', `bulk\_seeds', `sampling\_seed', `transform\_seed'\}
  
\setlength{\leftskip}{0.66cm}
    - default: every sampling receives a common set of entropy\_seeds per user specification which are shuffled and passed to each call
    
    - bulk\_seeds: every sampling receives a unique supplemental seed for every sampled entry for sampling from sampling\_generator (expended seed counts dependent on train/test/both configuration and numbers of rows). This scenario also defaults to sampling\_dict[`seeding\_type'] = `primary\_seeds'
    
    - sampling\_seed: every sampling operation receives one supplemental seed for sampling from sampling\_generator (expended seed counts dependent on train/test/both configuration)
    
    - transform\_seed: every noise transform receives one supplemental seed for sampling from sampling\_generator (expended seed counts are the same independent of train/test/both configuration)

\setlength{\leftskip}{0cm}

  - sampling\_dict[`\textbf{seeding\_type}'] defaults to `supplemental\_seeds' or `primary\_seeds' as described below, where `supplemental\_seeds' means that entropy seeds are integrated into np.random.SeedSequence with entropy seeding from the operating system. Also accepts `primary\_seeds', in which user passed entropy seeds are the only source of seeding. Please note that `primary\_seeds' is used as the default for the bulk\_seeds sampling\_type and `supplemental\_seeds' is used as the default for other sampling\_type options.

  - sampling\_dict[`\textbf{sampling\_report\_dict}'] defaults as False, accepts a prior populated postprocess\_dict[`sampling\_report\_dict'] from an automunge(.), call if this is not received it will be generated internally. sampling\_report\_dict is a resource for determining how many entropy\_seeds are needed for various sampling\_type scenarios.
  
  - sampling\_dict[`\textbf{stochastic\_count\_safety\_factor}']: defaults to 0.15, accepts float 0-1, is associated with the bulk\_seeds sampling\_type case and is used as a multiplier for number of seeds populated for sampling operations with a stochastic number of entries.
  
  - sampling\_dict[`\textbf{sampling\_generator}']: used to specify which generator will be used for sampling operations other than generation of additional entropy\_seeds. Defaults to `custom' (meaning the passed random\_generator or when unspecified the default PCG64), and accepts one of \{`custom', `PCG64', `MersenneTwister'\}.
  
  - sampling\_dict[`\textbf{extra\_seed\_generator}']: used to specify which generator will be used to sample additional entropy\_seeds when more are needed to meet requirements of sampling\_report\_dict, defaults to `custom' (meaning the passed random\_generator or when unspecified the default PCG64), and accepts one of \{`custom', `PCG64', `MersenneTwister', `off', `sampling\_generator'\}, where sampling\_generator matches specification for sampling\_generator, and `off' turns off sampling of additional entropy seeds.

* \textbf{noise\_augment}: accepts type int or float(int) \(\geq0\), defaults to 0. Used to specify 
a count of additional duplicates of training data prepared and concatenated with the
original train set. Intended for use in conjunction with noise injection, such that
the increased size of training corpus can be a form of data augmentation. (Noise injection
still needs to be assigned, e.g. by assigning root categories in assigncat or could
turn on automated noise with powertransform = `DP1'). Note that 
injected noise will be uniquely randomly sampled with each duplicate. When noise\_augment
is received as a dtype of int, one of the duplicates will be prepared without noise. When
noise\_augment is received as a dtype of float(int), all of the duplicates will be prepared 
with noise. When shuffletrain is activated the duplicates are collectively shuffled, and can distinguish
between duplicates by the original df\_train.shape in comparison to the ID set's Automunge\_index.
Please be aware that with large dataframes a large duplicate count may run into memory constraints,
in which case additional duplicates can be prepared separately in postmunge(.). Note that the entropy seed budget only accounts for preparing one set of data, for the noise\_augment option with entropy seeding we recommend passing a custom extra\_seed\_generator with a sampling\_type specification, which will result in internal samplings of additional entropy seeds for each additional noise\_augment duplicate (or for the bulk\_seeds case with external sampling can increase entropy\_seed budget proportional to the number of additional duplicates with noise).

\hyperref[A]{Table of Contents}

\newpage
\section{Noise injection tutorial}
\label{H}

\subsection{DP transformation categories}
\label{H1}

DP family of transforms are surveyed in the read me's library of transformations section as Differential Privacy Noise Injections. The noise injections can be performed in conjunction with numeric normalizations or categoric encodings, which options were surveyed in [Appendix \ref{C}] [Table \ref{root_cat_table}].

Here is an example of assigning some of these root categories to received features with headers `column1', `column2', `column3'. DTnb is z-score normalization with Gaussian noise to test data, shown here assigned to column1. DBod is ordinal encoding with weighted activation flips to both train and test data, shown here assigned to column2 and column3. (To just inject to train data the identifier string for that default configuration replaces the DT or DB prefix with DP.)
\begin{lstlisting}
assigncat = {'DTnb' : 'column1',
                 'DBod' : ['column2', 'column3']}
\end{lstlisting}

\hyperref[A]{Table of Contents}


\subsection{Parameter assignment}
\label{H2}

Each of these transformations accepts optional parameter specifications to vary from the defaults. Parameters are passed to transformations through the automunge(.) assignparam parameter. As we described in Appendix D, parameter assignments through assignparam can be conducted in three ways, where global\_assignparam passes the setting to every transform applied to every column, default\_assignparam pass the same setting to every instance of a specific transformation's tree category identifier applied to any column, or in the third option a parameter setting can be assigned to a specific transformation tree category identifier passed to a specific column (where that column may be an input column or a derived column with suffix appender passed to the transform). Note that the difference between a tree category and a root category \cite{anonymous_github} is that a root category is the identifier of the family tree of transformation categories assigned to a column in the assigncat parameter, and a tree category is an entry to one of those family tree primitives which is used to access the transformation function. To restate for clarity, the (column) string designates one of either the input column header (before suffixes are applied) or an intermediate column header with suffixes that serves as input to the target transformation.

\begin{lstlisting}
assignparam = {
  'global_assignparam'  : {'(parameter)': 42},
  'default_assignparam' : {'(category)' : {'(parameter)' : 42}},
  '(category)' : {'(column)'   : {'(parameter)' : 42}}}
\end{lstlisting}

For noise injections that are performed in conjunction with a normalization or encoding, the noise transform is generally applied in a different tree category than the encoding transform, so if parameters are desired to be passed to the encoding, assignparam will need to target a different tree category for the encoding than for the noise. Generally speaking, the noise transform family trees have been configured so that the noise tree category matches the root category, which was intentional for simplicity of parameter assignment (with an exception for DPhs for esoteric reasons). To view the full family tree such as to inspect the encoding tree category, the set of family trees associated with various root categories are provided in the code repository as FamilyTrees.md.

Note that assignparam can also be used to deviate from the default train or test noise injection settings. As noted above, the convention for the string identifiers of noise root categories is that `DP' injects noise to train and not test data, `DT' injects noise to test and not train data, and `DB' injects noise to both train and test data. These are the defaults, but each of these can be updated by parameter assignment with assignparam specification of `trainnoise' or `testnoise' parameters. 

As noted in [Appendix \ref{B}], for subsequent data passed to postmunge(.), the data can also be treated as test data or train data, and in both cases also have noise deactivated. The postmunge(.) traindata parameter defaults to False to prepare postmunge(.) as test data and accepts entries of \{False, True, `test\_no\_noise', `train\_no\_noise'\}.

Most of the noise injection transforms share common parameters between those targeting numeric or categoric entries.

\hyperref[A]{Table of Contents}

\subsection{Numeric parameters}
\label{H3}

\begin{itemize}

\item \textbf{trainnoise}: activates noise injection to train data (defaults True for DP or DB and False for DT)
\item \textbf{testnoise}: activates noise injection to test data (defaults True for DT or DB and False for DP)
\item \textbf{flip\_prob}: ratio of train entries receiving injection
\item \textbf{test\_flip\_prob}: ratio of test entries receiving injection (defaults as matched to flip\_prob)
\item \textbf{sigma}: scale of train noise distribution
\item \textbf{test\_sigma}: scale of test noise distribution
\item \textbf{mu}: mean of train noise distribution (before any scaling)
\item \textbf{test\_mu}: mean of test noise distribution (before any scaling)
\item \textbf{noisedistribution}: train noise distribution, defaults to `normal' (Gaussian), accepts one of {`normal', `laplace', `uniform', `abs\_normal', `abs\_laplace', `abs\_uniform', `negabs\_normal', `negabs\_laplace', `negabs\_uniform'}, where abs refers to all positive signed noise and negabs refers to all negative signed noise
\item \textbf{test\_noisedistribution}:  test noise distribution, comparable options supported
\item \textbf{rescale\_sigmas}: for min-max normalization (DPmm) or retain normalization (DPrt), this activates the mean adjustment noted in Appendix E, defaults to True
\item \textbf{retain\_basis}: for cases where distribution parameters passed as list or distribution, activating retain\_basis means the basis sampled in automunge is carried through to postmunge or the default of False means a unique basis is sampled in automunge and postmunge
\item \textbf{protected\_feature}: can be used to specify an adjacent categoric feature with sensitive attributes for a segment specific noise scaling adjustment which we speculate may reduce loss discrepancy
\end{itemize}
Default numeric parameters detailed in [Table \ref{defaultnumericsettings}]

\begin{table}[!ht]
    \centering
    \caption{Default numeric parameter settings}
    \begin{tabular}{|l|l|l|l|l|}
    \hline
          & DPnb & DPmm & DPrt & DPne \\ \hline
          & z-score & min-max & retain & pass-through \\ \hline
        flip\_prob & 0.03 & 0.03 & 0.03 & 0.03  \\ \hline
        test\_flip\_prob & flip\_prob & flip\_prob & flip\_prob & flip\_prob  \\ \hline
        sigma & 0.06 & 0.03 & 0.03 & 0.06  \\ \hline
        test\_sigma & 0.03 & 0.02 & 0.02 & 0.03  \\ \hline
        mu & 0 & 0 & 0  & 0 \\ \hline
        test\_mu & 0 & 0 & 0 & 0  \\ \hline
        noisedistribution & normal & normal & normal & normal  \\ \hline
        test\_noisedistribution & normal & normal & normal & normal  \\ \hline
        rescale\_sigmas  & - & True & True & -\\ \hline
        retain\_basis & False & False & False & False \\ \hline
        protected\_feature & False & False & False & False \\ \hline
    \end{tabular}
    \label{defaultnumericsettings}
\end{table}

\hyperref[A]{Table of Contents}

\subsection{Categoric parameters}
\label{H4}

\begin{itemize}

\item \textbf{trainnoise}: activates noise injection to train data (defaults True for DP or DB and False for DT)
\item \textbf{testnoise}: activates noise injection to test data (defaults True for DT or DB and False for DP)
\item \textbf{flip\_prob}: ratio of train entries receiving injection
\item \textbf{test\_flip\_prob}: ratio of test entries receiving injection
\item \textbf{weighted}: weighted vs uniform sampling of activation flips to train data
\item \textbf{test\_weighted}: weighted vs uniform sampling of activation flips to test data
\item \textbf{retain\_basis}: for cases where distribution parameters passed as list or distribution, activating retain\_basis means the basis sampled in automunge is carried through to postmunge or the default of False means a unique basis is sampled in automunge and postmunge
\item \textbf{protected\_feature}: can be used to specify an adjacent categoric feature with sensitive attributes for a segment specific noise scaling adjustment which we speculate may reduce loss discrepancy
\end{itemize}

Default categoric parameters detailed in [Table \ref{defaultcategoricsettings}]

\begin{table}[!ht]
    \centering
    \caption{Default categoric parameter settings}
    \begin{tabular}{|l|l|l|l|l|l|l|l|l|l|}
    \hline
        ~ & DPbn & DPod & DP10 & DPoh & DPhs & DPh2 & DPse & DPsk & DSPpc \\ \hline
        ~ & boolean & ordinal & binarized & one hot & multi hash & hash & swap & mask & pass-through  \\ \hline
        flip\_prob & 0.03 & 0.03 & 0.03 & 0.03 & 0.03 & 0.03 & 0.03 & 0.03 & 0.03 \\ \hline
        test\_flip\_prob & 0.01 & 0.01 & 0.01 & 0.01 & 0.01 & 0.01 & 0.01 & 0.01 & 0.01 \\ \hline
        weighted & True & True & True & True & True & True & - & - & True \\ \hline
        test\_weighted & True & True & True & True & True & True & - & - & True \\ \hline
        swap\_noise & - & - & False & False & - & - & - & - & - \\ \hline
        retain\_basis & False & False & False & False & False & False & False & False & False \\ \hline
        protected\_feature & False & False & False & False & False & False & - & - & False \\ \hline
        mask\_value &- & - & - & - & - & - & - & 0 & - \\ \hline
    \end{tabular}
    \label{defaultcategoricsettings}
\end{table}

Here is an example of assignparam specification to: 

\begin{itemize}

\item set an all positive noise distribution for category DPmm as applied to an input column with header `column1', noting that for scaled noise like DPmm all positive or all negative noise should be performed with a deactivated noise\_scaling\_bias\_offset.
\item update the flip\_prob parameter to 0.1 for all cases of DPnb injections via default\_assignparam
\item apply testnoise injections to all noise transforms via global\_assignparam

\end{itemize}

\begin{lstlisting}
#assumes DPmm and DPnb have been assigned in assigncat
assignparam = {
  'DPmm' : {'inputcolumn': {'noisedistribution'         : 'abs_normal',
                                     'noise_scaling_bias_offset' : False}},
  'default_assignparam'  : {'DPnb' : {'flip_prob' : 0.1}},
  'global_assignparam'    : {'testnoise': True},
}
\end{lstlisting}

\hyperref[A]{Table of Contents}


\subsection{Noise injection under automation}
\label{H5}

The automunge(.) powertransform parameter can be used to select between alternate sets of default transformations applied under automation. We currently have six scenarios for default encodings with noise, including powertransform passed as one of {`DP1', `DP2'}. DP2 differs from DP1 in that numerical defaults to retain normalization instead of z-score and categoric defaults to ordinal instead of binarization. (DT and DB equivalents DT1 / DT2 / DB1 / DB2 allow for different default train and test noise configurations, i.e. DT injects just to test data and DB to both train and test [Appendix \ref{B}].)

Shown following are the root categories applied under automation for these two powertransform scenarios of `DP1' or `DP2'.

powertransform = `DP1'
\begin{itemize}

\item numeric receives DPnb
\item categoric receives DP10
\item binary receives DPbn
\item hash receives DPhs
\item hsh2 receives DPh2 
\item (labels do not receive noise)

\end{itemize}

powertransform = `DP2'
\begin{itemize}

\item numeric receives DPrt
\item categoric receives DPod
\item binary receives DPbn
\item hash receives DPhs
\item hsh2 receives DPh2 
\item (labels do not receive noise)

\end{itemize}

Otherwise noise can just be manually assigned in the assigncat parameter as demonstrated above, which specifications will take precedence over what would otherwise be performed under automation.


\hyperref[A]{Table of Contents}

\subsection{Data augmentation with noise}
\label{H6}

Data augmentation refers to increasing the size of a training set with manipulations to increase variety. In the image modality it is common to achieve data augmentation by way of adjustments like image cropping, rotations, color shift, etc. Here we are simply injecting noise to training data for similar effect. In a deep learning benchmark performed in \cite{anonymous_Numbers} it was found that this type of data augmentation was fairly benign with a fully represented data set, but was increasingly beneficial with underserved training data. Note that this type of data augmentation can be performed in conjunction with non-deterministic inference by simply injecting to both train and test data.

Data augmentation can be realized by assigning noise transforms in conjunction with the automunge(.) noise\_augment parameter, which accepts integers of number of additional duplicates to prepare, e.g. noise\_augment=1 would double the size of the training set returned from automunge(.). For cases where too much duplication starts to run into memory constraints additional duplicates can also be prepared with postmunge(.), which also has a noise\_augment parameter option and accepts the traindata parameter to distinguish whether a data set is to be treated as train or test data.

Under the default configuration when noise\_augment is received as an integer dtype, one of the duplicates will be prepared without noise. If noise\_augment is received as a float(int) type, all of the duplicates will be prepared with noise.

Here is an example of preparing data augmentation for the data set loaded earlier.

\begin{lstlisting}
train, train_ID, labels, \
val, val_ID, val_labels, \
test, test_ID, test_labels, \
postprocess_dict = \
am.automunge(df_train,
                 powertransform = 'DP2',
                 noise_augment = 2.0,
                 printstatus = False)
\end{lstlisting}


\hyperref[A]{Table of Contents}

\subsection{Alternate random samplers}
\label{H7}

The random sampling for noise injection defaults to numpy's PCG64, which is based on the PCG pseudo random number generator algorithm \cite{ONeill2014PCGA}. On its own this generator is not truly random, it relies on seedings of entropy provided by the operating system which are then enhanced through use. To support integration of enhanced randomness profiles, both automunge(.) and postmunge(.) accept parameters for entropy\_seeds and random\_generator.

\textbf{entropy\_seeds} accepts an integer or list/array of integers which may serve as a supplemental source of entropy for the numpy.random generator to enhance randomness properties.

\textbf{random\_generator} accepts input of a numpy.random.Generator formatted random sampler. An example could be numpy.random.MT19937 for Mersenne Twister, or could even be an external library with a numpy.random formatted generator, such as for example could be used to sample with the support of quantum circuits.

Specifications of entropy\_seeds and random\_generator are specific to an automunge(.) or postmunge(.) call, in other words they are not returned in the populated postprocess\_dict. The two parameters can also be passed in tangent, for sampling with a custom generator with custom supplemental entropy seeds. 

If an alternate library does not have a numpy.random formatted generator, their output can be channeled to entropy\_seeds for similar benefit. Here is an example of specifying an alternate generator and supplemental entropy seedings.

\begin{lstlisting}
random_generator = numpy.random.MT19937
entropy_seeds = [4,5,6]
\end{lstlisting}

In the default case the same bank of entropy seeds is fed to each sampling operation with a shuffle. The library also supports different types of sampling scenarios that can be supported by entropy seedings. Alternate sampling scenarios can be specified to automunge(.) or postmunge(.) by the sampling\_dict parameter. Here are a few scenarios to illustrate.

1) In one scenario, instead of passing externally sampled supplemental entropy seeds, a user can pass a custom generator for internal sampling of entropy seeds. Here is an example of using a custom generator to sampling entropy seeds and the default generator PCG64 for sampling applied in the transformations. The sampling\_type bulk\_seeds means that a unique seed will be generated for each sampled entry. When not sampling externally, this scenario may be beneficial for improving utilization rate of quantum hardware since the quantum sampling will only take once per automunge(.) or postmunge(.) call and latency will be governed by the sampler instead of pandas operations.
\begin{lstlisting}
random_generator = (custom numpy formatted generator)
entropy_seeds = False
sampling_dict = \
{'sampling_type' : 'bulk_seeds',
 'extra_seed_generator' : 'custom',
 'sampling_generator' : 'PCG64',
 }
\end{lstlisting}

2) In another scenario a user may want to reduce their sampling budget by only accessing one entropy seed for each set of entries. This is accessed with the sampling\_type of sampling\_seed.
\begin{lstlisting}
random_generator = (custom numpy formatted generator)
entropy_seeds = False
sampling_dict = \
{'sampling_type' : 'sampling_seed',
 'extra_seed_generator' : 'custom',
 'sampling_generator' : 'PCG64',
 }
\end{lstlisting}

3) There may be a case where a source of supplemental entropy seeds isn't available as a numpy.random formatted generator. In this case, in order to apply one of the alternate sampling\_type scenarios, a user may desire to know a budget of how many seeds are required for externally sampled seeds passed through the entropy\_seeds parameter. This can be accomplished by first running the automunge(.) call without entropy seeding specifications to generate the report returned as postprocess\_dict[`sampling\_report\_dict']. (note that if sampling seeds internally with a custom generator this isn't needed.) Note that the sampling\_report\_dict will report requirements separately for train and test data and in the bulk\_seeds case will have a row count basis. (If not passing test data to automunge(.) the test budget can be omitted. For postmunge the use of train or test budget should align with the postmunge traindata parameter.) For example, if a user wishes to derive a set of entropy seeds to support a bulk\_seeds sampling type, they can produce a report and derive as follows:
\begin{lstlisting}
#first run automunge(.) to populate postprocess_dict (not shown)
#using comparable category and parameter assignments
#we recommend running initially with default sampling_type
#to populate sampling_report_dict for test data even if df_test not provided

#access the sampling_report_dict in the returned postprocess_dict
sampling_report_dict = postprocess_dict['sampling_report_dict']

#a bulk_seeds sampling_type budget will need to take account for row counts
rowcount_train = df_train.shape[0]
rowcount_test = df_test.shape[0]

#the budget can be derived as
train_budget = \
sampling_report_dict['bulk_seeds_total_train'] * rowcount_train \
/ sampling_report_dict['rowcount_basis_train']

test_budget = \
sampling_report_dict['bulk_seeds_total_test'] * rowcount_test \
/ sampling_report_dict['rowcount_basis_test']

#number of external seeds needed for bulk seeds case:
seed_count = \
train_budget + test_budget

#this number of seeds can then be passed to the entropy_seeds parameter

random_generator = False
entropy_seeds = externally_sampled_seeds_list
sampling_dict = \
{'sampling_type' : 'bulk_seeds',
 'extra_seed_generator' : 'PCG64',
 'sampling_generator' : 'PCG64',
 }
\end{lstlisting}

\hyperref[A]{Table of Contents}


\subsection{QRAND library sampling}
\label{H8}

To sample noise from a quantum circuit, a user can either pass externally sampled entropy\_seeds or make use of an external library with a numpy.random formatted generator. Here's an example of using the QRAND library \cite{QRAND_library} to sample from a quantum circuit, based on a tutorial provided in their read me which makes use of Qiskit \cite{Qiskit}.

\begin{lstlisting}
from qrand import QuantumBitGenerator
from qrand.platforms import QiskitPlatform
from qrand.protocols import HadamardProtocol
from qiskit import IBMQ

provider = IBMQ.load_account()
platform = QiskitPlatform(provider)
protocol = HadamardProtocol()
bitgen = QuantumBitGenerator(platform, protocol)

#then can initialize automunge(.) or postmunge(.) parameters
#for each sampling being channeled through the quantum circuit
random_generator = bitgen
entropy_seeds = False
sampling_dict = \
{'sampling_type' : 'default',
 'extra_seed_generator' : 'off',
 'sampling_generator' : 'custom',
 }
 
#or if you only want to access the quantum circuit once per data set
#can initialize in this alternate configuration for similar result
random_generator = bitgen
entropy_seeds = False
sampling_dict = \
{'sampling_type' : 'bulk_seeds',
 'extra_seed_generator' : 'custom',
 'sampling_generator' : 'PCG64',
 }
\end{lstlisting}

\hyperref[A]{Table of Contents}


\subsection{All together now}
\label{H9}

Let's do a quick demonstration tying it all together. Here we'll apply the powertransform = `DP2' option for noise under automation, override a few of the default transforms with assigncat, assign a few deviations to transformation parameters via assignparam, add some additional entropy seeds from some other resource, and prepare a few additional training data duplicates for data augmentation purposes.

\begin{lstlisting}
powertransform = 'DP2'
assigncat = {'DPh2' : 'Name'}
noise_augment = 2.
entropy_seeds = [432,6,243,561232,89]

#(Age is a feature header in the Titanic data set)
assignparam = {
    'DPrt' : {'Age': {'noisedistribution'         : 'abs_normal',
                            'noise_scaling_bias_offset' : False}},
    'default_assignparam' : {'DPrt' : {'flip_prob' : 0.1}},
    'global_assignparam'  : {'testnoise': True}}

train, train_ID, labels, \
val, val_ID, val_labels, \
test, test_ID, test_labels, \
postprocess_dict = \
am.automunge(df_train,
                 labels_column = labels_column,
                 trainID_column = trainID_column,
                 powertransform = powertransform,
                 assigncat = assigncat,
                 noise_augment = noise_augment,
                 entropy_seeds = entropy_seeds,
                 assignparam = assignparam,
                 printstatus = False)
\end{lstlisting}

Similarly we can prepare additional test data in postmunge(.) using the postprocess\_dict returned from automunge(.), which since we set testnoise as globally activated will result in injected noise in the default traindata=False case.

\begin{lstlisting}
entropy_seeds = [2345, 77887, 2342, 7878789]

traindata = False

test, test_ID, test_labels, \
postreports_dict = \
am.postmunge(postprocess_dict, 
                 df_test,
                 entropy_seeds = entropy_seeds,
                 traindata=traindata,
                 printstatus=False)
\end{lstlisting}

\hyperref[A]{Table of Contents}


\subsection{Noise directed at existing data pipelines}
\label{H10}

One more thing. When noise is intended for direction at an existing data pipeline, such as for incorporation of noise into test data for an inference operation on a previously trained model, there may be desire to inject noise without other edits to a dataframe. This is possible by passing the dataframe as a df\_train to an automunge(.) call to populate a postprocess\_dict with assignment of the various features to one of these four pass-through categories:

\begin{itemize}
\item DPne: pass-through numeric with gaussian (or laplace) noise, comparable parameter support to DPnb
\item DPse: pass-through with swap noise (e.g. for categoric data), comparable parameter support to DPmc
\item DPpc: pass-through with weighted categoric noise (categoric activation flips), comparable parameter support to DPod
\item excl: pass-through without noise
\end{itemize}

Once populated, the postprocess\_dict can be used to prepare additional data in postmunge(.) which has lower latency. Note that DPse injects swap noise by accessing an alternate row entry for a target. This type of noise may not be suitable for test data injections in a scenario where inference may be run on a test set with one or very few samples. The convention in library is that data is received in a tidy form (one column per feature and one row per observation), so ideally categoric features should be received in a single column configuration for targeting with DPse.

Note that DPne will return entries as float data type, converting any non-numeric to NaN. The default noise scale for DPne (sigma=0.06 / test\_sigma=0.03) is set to align with z-score normalized data. For the DPne pass-through transform, since the feature may not be z-score normalized, the scaling is adjusted by multiplication with the evaluated standard deviation of the feature as found in the training data by use of the defaulted parameter rescale\_sigmas = True. This adjustment factor is derived based on the training data used to fit the postprocess\_dict, and that same basis is carried through to postmunge(.). If user doesn't have access to the training data, they can fit the postprocess\_dict to a representative df\_test routed as the automunge(.) df\_train.

Having populated the postprocess\_dict, additional inference data can be channeled through postmunge, which has latency benefits.

\begin{lstlisting}
numeric_features = ['Age', 'Fare']

categoric_features = \
['Pclass', 'Name', 'Sex', 'SibSp', 'Parch', 'Ticket', 'Cabin', 'Embarked']

passthrough_features = ['PassengerId']

#assign the features to one of DTne/DTse/excl
#The DT configuration defaults to injecting noise just to test data
assigncat = \
{'DTne' : numeric_features, #numeric features receiving gaussian noise
 'DTse' : categoric_features, #categoric features receiving swap noise
 'excl' : passthrough_features,
}

#if we want to update the noise parameters they can be applied in assignparam
#shown here are the defaults
assignparam = \
{'default_assignparam' : 
  {'DPne' : {'test_sigma' : 0.06,
                 'rescale_sigmas' : True},
   'DTse' : {'test_flip_prob' : 0.01}}}

#We'll also deactivate shuffletrain to retain order of rows
shuffletrain = False

#note that the family trees for DPne / DPse / DPpc / excl 
#do not include NArw aggregation
#so no need to deactivate NArw_marker
#they are also already excluded from infill based on process_dict specification
#so no need to deactivate MLinfill

#the orig_headers parameter retains original column headers 
#without suffix appenders
orig_headers = True

#this operation can fit the postprocess_dict to the df_test (or df_train)
train, train_ID, labels, \
val, val_ID, val_labels, \
test, test_ID, test_labels, \
postprocess_dict = \
am.automunge(df_test,
                 assigncat = assigncat,
                 assignparam = assignparam,
                 shuffletrain = shuffletrain,
                 orig_headers = orig_headers,
                 printstatus = False)
            
#we can then use the populated postprocess_dict to run postmunge(.)
#which has better latency than automunge(.)
#the entropy seeding parameters are shown with their defaults for reference

test, test_ID, test_labels, \
postreports_dict = \
am.postmunge(postprocess_dict, 
                 df_test,
                 printstatus = False,
                 random_generator = False,
                 entropy_seeds = False,
                 sampling_dict = {}
                 )
\end{lstlisting}

The returned dataframe test can then be passed to inference. The order of columns in returned dataframe will be retained for these transforms and the orig\_headers parameter retains original column headers without suffix appenders.

The postmunge(.) call can then be repeated as additional inference data becomes available, and could be applied sequentially to streams of data in inference.

\hyperref[A]{Table of Contents}

\newpage
\onecolumn
\section{Advanced noise profiles}
\label{Ia}

The noise profiles discussed thus far have mostly been a composition of two perturbation vectors, one arising from a Bernoulli sampling to select injection targets and a second arising from either a distribution sampling for numeric or a choice sampling for categoric. There may be use cases where a user desires some additional variations in the form of distribution. This section will survey a few more advanced noise profile compositions methods available. Compositions beyond those discussed here are also available by custom defined transformation functions which are available by use of a simple template.

In some sense, the methods discussed here will be a form of probabilistic programming, although not a Turing complete one. For Turing complete distribution compositionally, we recommend channeling through custom defined transformation functions that make use of external libraries with such capability, e.g. \cite{tran2017edward}. Custom transformations can apply a simple template detailed in the read me \cite{anonymous_github} section ``Custom Transformation Functions."

\subsection{Noise parameter randomization}
\label{I1}

The generic numeric noise injection parameters were surveyed in [Appendix H.3] and their defaults presented in [Table \ref{defaultnumericsettings}]. Similarly, the generic categoric noise injection parameters were surveyed in [Appendix H.4] with defaults presented in [Table \ref{defaultcategoricsettings}]. For each of the parameters related to noise scaling, weighting, or specification, the library offers options to randomize their derivation by a random sampling, including support for such sampling to be conducted with the support of entropy seeding. The random sampling of parameter values can either be activated by passing the parameters as a list of candidate values for a choice sampling between, or for parameters with float specification by passing parameters as arbitrary scipy stats \cite{SciPy2020} formatted distributions for a shaped sampling. 

One of the parameters that we did not go into great detail in earlier discussions was the `retain\_basis' parameter. This parameter is relevant to noise parameter randomization, and refers to the practice of applying a common or unique noise parameter sampling between a feature as prepared in the test data received by automunge(.) and each subsequent postmunge(.) call. We expect that in most cases a user will desire a common noise profile between initial test data prepared in automunge(.) and subsequent test data prepared in postmunge(.) for inference, as is the default True setting. A consistent noise profile should be appropriate when relying on a corresponding noise profile injected to the training data. We speculate that there may be cases where non-deterministic inference could benefit from a unique sampled noise profile across inference operations. Deactivating the retain\_basis option can either be conducted specific to a feature, or may be conducted globally using an assignparam[`global\_assignparam'] specification.

\hyperref[A]{Table of Contents}

\subsection{Noise profile composition}
\label{I2}

Another channel for adding additional perturbation vectors into a noise profile is available by composing sets of noise injection transforms using the family tree primitives [Table \ref{sample-table}]. The primitives are for purposes of specifying the order, type, and retention of derivations applied when a `root category' is assigned to an input feature, where each derivation is associated with a `tree category' populated in either the root category's family or some downstream family tree accessed from origination of the root category. As they are implemented with elements of recursion, they can be used to specify transformation sets that include generations and branches of univariate derivations. Thus, multiple noise injection operations can be applied to a single returned set, potentially including noise of different scaling and/or injection ratios.

Here is an example of family tree specification for a numeric injection of Gaussian noises with two different profiles as applied downstream of a z-score normalization.

\begin{lstlisting}
transformdict = {}
transformdict.update({'DPnb' : {'parents'       : ['DPn3'],
                                          'siblings'      : [],
                                          'auntsuncles'   : [],
                                          'cousins'       : ['NArw'],
                                          'children'      : [],
                                          'niecesnephews' : [],
                                          'coworkers'     : ['DPnb2'],
                                          'friends'       : []}})

transformdict.update({'DPn3' : {'parents'       : ['DPn3'],
                                          'siblings'      : [],
                                          'auntsuncles'   : [],
                                          'cousins'       : [],
                                          'children'      : ['DPnb'],
                                          'niecesnephews' : [],
                                          'coworkers'     : [],
                                          'friends'       : []}})

processdict = {}
processdict.update({'DPnb' : {'functionpointer' : 'DPnb',
                                        'defaultparams' : {'sigma':0.5,
                                                                 'flip_prob':0.0001}}})
                                                 
processdict.update({'DPnb2' : {'functionpointer' : 'DPnb',
                                         'defaultparams' : {'sigma':0.05,
                                                                  'flip_prob':0.03}}})
                                                                  
processdict.update({'DPn3' : {'functionpointer' : 'nmbr'}})
\end{lstlisting}

\hyperref[A]{Table of Contents}

\subsection{Protected attributes}
\label{Protectedattributes}

We noted in our related work discussions in Section 7 that one possible consequence of noise injections is that different segments of a feature set corresponding to categories in an adjacent protected feature may be impacted more than others owing to a diversity in segment distributions in comparison to a common noise profile, which may contribute to loss discrepancy between categories of that protected feature \cite{pmlr-v119-khani20a} without mitigation. The mitigation available from Automunge was inspired by the description of loss discrepancy offered by the citation, and might be considered another contribution of this paper. 

Loss discrepancy may arise in conjunction with noise due to the property that segments of a feature that are not randomly sampled in some cases may not share the same distribution profile from their aggregation. In some cases the segments in a noise targeted feature corresponding to the attributes of an adjacent protected features may have this property. Thus by injecting a single noise profile into segments with different scalings, those segments will be unequally impacted.

The Automunge solution is an as yet untested hypothesis with a clean implementation. When a user specifies an adjacent protected feature for a numeric noise feature, the noise scaling for each segment of the noise target feature corresponding to attributes in the adjacent feature is rescaled, with the train data basis carried through to test data. For example, if the aggregate feature has standard deviation \(\sigma_a\), and the segment has standarde deviation \(\sigma_s\), the noise can be scaled for that segment by multiplication by the ratio \(\sigma_s\ / \sigma_a\). Similarly, if a protected feature is specified for a categoric noise feature, the derivation of weights by frequency counts can be calculated for each segment individually. In both cases, the segment noise distributions will share a common profile between train and test data, and the aggregate noise distribution will too as long as the distribution properties of the protected feature remain consistent. These options can be activated by passing an input header string to the protected\_feature parameter of a distribution sampled or weighted categoric noise transform through assignparam, with support for up to one protected feature per noise targeted feature.

\hyperref[A]{Table of Contents}

\newpage
\section{Distribution scaling}
\label{E}

The paper noted in Section 3.1 that the numeric injections may have the potential to increase the maximum value found in the returned feature set or decrease the minimum value. In some cases, applications may benefit from retention of such feature properties before and after injection. When a feature set is scaled with a normalization that produces a known range of values, as is the case with min-max scaling (which scales data to the range between 0–1 inclusive), it becomes possible to manipulate noise as a function of entry properties to ensure retained range of values after injection. Other normalizations with a known range other than 0–1 (such as for `retain’ normalization \cite{anonymous_Numbers}) can be shifted to the range 0–1 prior to injection and then reverted after for comparable effect. As this results in noise distribution derived as a function of the feature distribution, the sampled noise mean can be adjusted to closer approximate a zero mean for the scaled noise.

The noise scaling [Alg \ref{alg:noisescaling}] assumes a min-max scaled feature, and thus this algorithm works because both min-max and noise have a known range of entries. The pandas implementation looks a little different, this is equivalent.

\begin{algorithm}[tbh]
   \caption{Noise scaling}
   \label{alg:noisescaling}
\begin{algorithmic}
\STATE \#cap noise outliers:
   \IF{(noise) $>$ 0.5} 
   \STATE (noise) = 0.5
   \ELSIF{(noise) $<$ -0.5} 
   \STATE (noise) = -0.5
   \ENDIF
\STATE \#scale noise based on min-max entry properties:
   \IF{(minmax) $<$ 0.5} 
   \IF{(noise) $<$ 0}
   \STATE (scaled noise) = (noise) * (minmax) / 0.5
   \ELSIF{(noise) $\geq$ 0}
   \STATE (scaled noise) = (noise)
   \ENDIF
   \ELSIF{(minmax) $\geq$ 0.5}
   \IF{(noise) $<$ 0}
   \STATE (scaled noise) = (noise)
   \ELSIF{(noise) $\geq$ 0}
   \STATE (scaled noise) = (noise) * (1 - (minmax)) / 0.5
   
   \ENDIF
   \ENDIF

\end{algorithmic}
\end{algorithm}

Since the noise scaling is based on the feature distribution which is unknown, the mean adjustment to closer approximate a zero mean after scaling is derived by linear interpolation from iterated scalings to derive a final scaling used for seeded sampling and injection [Alg \ref{alg:noisescaling_mean_adjustment}]. Note that this final adjusted mean \(\mu_{adjusted}\) is derived based on the training data and that same basis is used for subsequent test data. 

\begin{algorithm}[tbh]
   \caption{Noise scaling mean adjustment and injection}
   \label{alg:noisescaling_mean_adjustment}
\begin{algorithmic}
\STATE \(\mu_{0}\) = known noise mean before scaling 
\STATE \(\mu_{1}\) = mean of noise of mean \(\mu_{0}\) scaled with [Alg \ref{alg:noisescaling}]
\STATE \(\mu_{2}\) = mean of noise of mean \(\mu_{1}\) scaled with [Alg \ref{alg:noisescaling}]
\STATE
\STATE \(\mu_{adjusted}\) = \(\mu_{0}\) + (\(\mu_{0}\) - \(\mu_{1}\)) / (\(\mu_{2}\) - \(\mu_{1}\)) * (\(-\mu_{1}\))
\STATE
\STATE (scaled noise) = sampled noise with mean \(\mu_{adjusted}\) and scaled with [Alg \ref{alg:noisescaling}]
\STATE
\STATE (minmax) + (Bernoulli) * (scaled noise) = (injected)

\end{algorithmic}
\end{algorithm}

\hyperref[A]{Table of Contents}

\newpage
\section{Causal inference}
\label{G}

As a related tangent, in causal inference \cite{pearl2019seven} an approach for identifying causal direction makes use of independent component analysis (ICA) \cite{HYVARINEN2000411}, which recovers independent components from a given signal \cite{richard2021adaptive}. \cite{JMLR:v7:shimizu06a} suggests that for linear ICA a distinction may be warranted between noise of different profiles, as Gaussian perturbation vectors are exceptions to causal inference's ability to detect direction of graph structures. In the context of stochastic perturbations, we suspect this might be a positive property of gaussian noise in comparison to other noise profiles, as it won’t influence directional derivations from other non gaussian vectors. This gaussian exception appears to vanish when non-linear ICA is performed with interventions for non-stationarity \cite{pmlr-v89-hyvarinen19a}. It should be noted that the output of non-deterministic inference will likely have a non-symmetric noise distribution of exotic nature.

The focus of the paper was non-deterministic inference by noise injection to tabular features in the context of a supervised learning workflow. Because the application of Automunge is intended to be applied directly preceding passing data to training and inference, we don't expect there will be a great risk of noise injections polluting data that would otherwise be passed to causal inference. However it should be noted that the output of non-deterministic inference will likely have a non-symmetric noise distribution of exotic nature based on properties of the model, and any downstream use of causal inference may be exposed to that profile. One way to circumvent this could be to run inference redundantly with and without noise such as to retain the deterministic inference outcome as a point of reference. Such operation can easily be performed by passing the same test data to the Automunge postmunge(.) function with variations on the traindata parameter as per [Appendix \ref{B}].



\hyperref[A]{Table of Contents}
\newpage
\twocolumn
\section{Benchmarking}
\label{Benchmarking}
A series of sensitivity analysis trials were conducted to evaluate different injection scenarios and parameter settings [Appendix \ref{I}, \ref{J}, \ref{K}]. As we were primarily interested in relative performance between configurations we narrowed focus to a single data set, a contributing factor was also resource limitations. Scenarios were trained on a dedicated workstation over the course of about two months. The IEEE-CIS data set \cite{Vesta_IEEE} was selected for scale and familiarity. The data was prepared in configurations of just numeric features, just categoric features, and all features. Injections were performed without supplemental entropy seeding. In addition to parameter setting for numeric or categoric noise types and scales, the benchmarks also considered scenarios associated with injections just to train data, just to test data, and both train and test data (which we refer to as the traintest scenario). The trials were repeated in a few learning frameworks appropriate for the tabular domain, including neural networks via the Fastai library \cite{Howard2020} (with four dense layers and width equal to feature count), cat boosting via the Catboost library \cite{dorogush2018catboost}, and gradient boosting with XGBoost \cite{XGBoost2016}, and in each case with default hyperparameters for consistency. The primary considerations of interest were validating default noise parameter settings [App \ref{H3}, \ref{H4}] (they were confirmed as conservative) and demonstrating noise sensitivity in context of relative performance between injections targeting train / test / traintest features. The full results are presented in [Appendix \ref{I}, \ref{J}, \ref{K}], and a representative set of jupyter notebooks used to aggregate these figures is included in the supplementary material.

\subsection{Neural Networks}

\begin{figure}[ht]
\vskip 0.2in
\begin{center}
\centerline{\includegraphics[width=0.9\columnwidth]{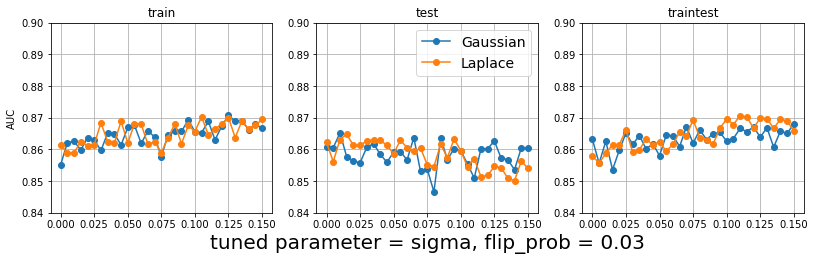}}

\centerline{\includegraphics[width=0.9\columnwidth]{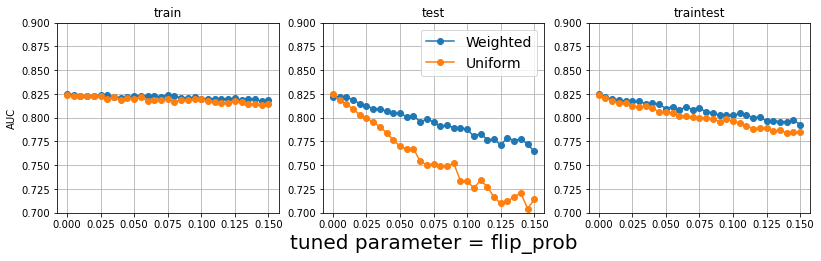}}
\caption{Fastai numeric and categoric injections}
\label{fastai_weighted_uniform}
\end{center}
\vskip -0.2in
\end{figure}

The neural network [Appendix \ref{I}] had a material performance benefit from injection of distribution sampled noise into training numeric features, which had the added benefit of making the model robust to corresponding noise injections into test features [Fig \ref{fastai_weighted_uniform}]. Categoric injections resembled a linear degradation of performance with increasing noise profile. There was a performance degradation from injections to test data in isolation, but when both were applied the performance profile was very similar between injecting just to training data verses injecting to both train and test. The exception was for numeric injections of swap noise, which had the linear degradation profile resembling the categoric noise. Neural networks were the only case demonstrating regularization from training noise injections. 

\begin{figure}[ht]
\vskip 0.2in
\begin{center}
\centerline{\includegraphics[width=0.9\columnwidth]{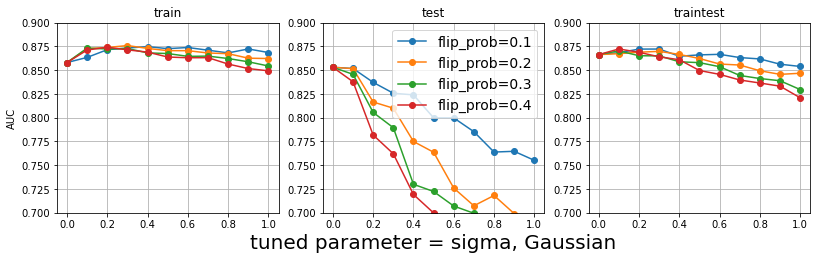}}

\caption{Fastai Gaussian scale with Bernoulli ratio scenarios}
\label{Fastai_Gaussian_scale}
\end{center}
\vskip -0.2in
\end{figure}

For numeric injections, there did not appear to be a noticeable difference in performance when comparing Gaussian and Laplace noise, suggesting that in non-deterministic inference the exposure to a wider range of inference scenarios from thicker tails in the Laplace noise distribution, as may be desirable in some applications, was neutral compared to thin tailed distributions. The performance sensitivity profile from increasing the numeric distribution scaling appeared very similar to the performance sensitivity profile from increasing ratio of injected entries by the Bernoulli sampling. An interesting finding was to what extent the preceding Bernoulli sampling (which resulted in only a sampled ratio of entries receiving noise injections) increased the model's tolerance for increase scale of noise distribution [Fig \ref{Fastai_Gaussian_scale}].

\subsection{Cat Boosting}

Cat boosting [Appendix \ref{J}] tolerated training data injections better than gradient boosting, and injections to both train and test data closely resembled injections to just test data. 

\subsection{Gradient Boosting}

\begin{figure}[ht]
\vskip 0.2in
\begin{center}
\centerline{\includegraphics[width=0.9\columnwidth]{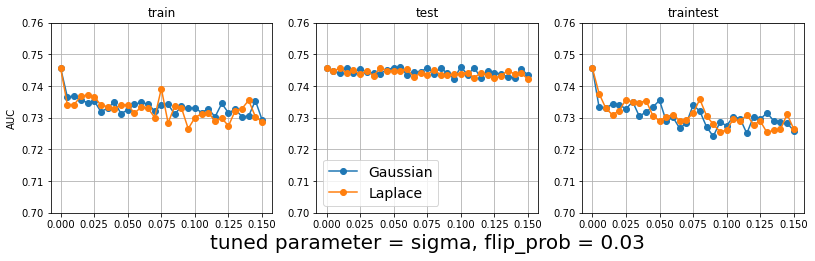}}

\centerline{\includegraphics[width=0.9\columnwidth]{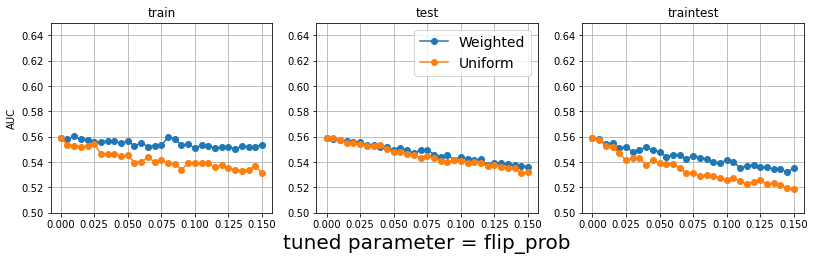}}
\caption{XGBoost numeric and categoric injections}
\label{xgboost_weighted_uniform}
\end{center}
\vskip -0.2in
\end{figure}

\begin{figure}[ht]
\vskip 0.2in
\begin{center}

\centerline{\includegraphics[width=0.9\columnwidth]{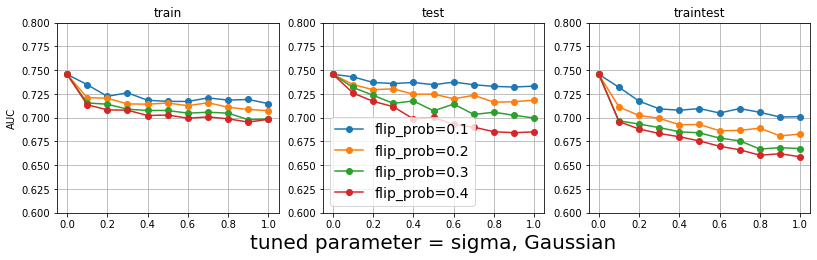}}

\caption{XGBoost Gaussian scale with Bernoulli ratio scenarios}
\label{XGBoost_Gaussian_scale}
\end{center}
\vskip -0.2in
\end{figure}

Gradient boosting [Appendix \ref{K}] demonstrated some material distinctions in comparison to the neural network [Fig \ref{xgboost_weighted_uniform}, \ref{XGBoost_Gaussian_scale}]. Instead of training data injections benefiting performance, it actually degraded, and the traintest scenario resembled the train scenario. The strongest performance came from injections only to the test features for inference, which for the numeric injections models were robust to mild noise profiles. This suggests that non-deterministic inference can be integrated directly into an existing gradient boosting pipeline with a manageable primary performance impact, which impact, without the regularizing effect from training available to neural networks, will be in play for gradient boosting in the tradeoffs between balancing a performance metric and realizing non-determinism.


\hyperref[A]{Table of Contents}

\newpage
\twocolumn
\section{Sensitivity analysis - Fastai}
\label{I}


\textbf{Highlights:}

\begin{figure}[ht]
\vskip 0.2in
\begin{center}
\centerline{\includegraphics[width=\columnwidth]{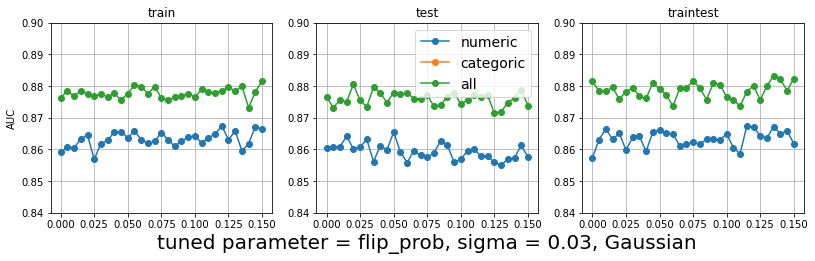}}

\centerline{\includegraphics[width=\columnwidth]{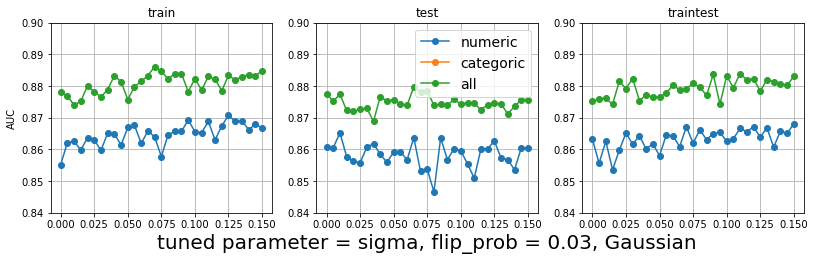}}
\caption{Fastai Gaussian injections}
\label{fastai_Gaussian_tuned_seperately}
\end{center}
\vskip -0.2in
\end{figure}

\begin{figure}[ht]
\vskip 0.2in
\begin{center}

\centerline{\includegraphics[width=\columnwidth]{Exp040.4numericflip_prob_0107b_varytogether_tuned_sigma_fastai_fliprobscenarios.png}}

\centerline{\includegraphics[width=\columnwidth]{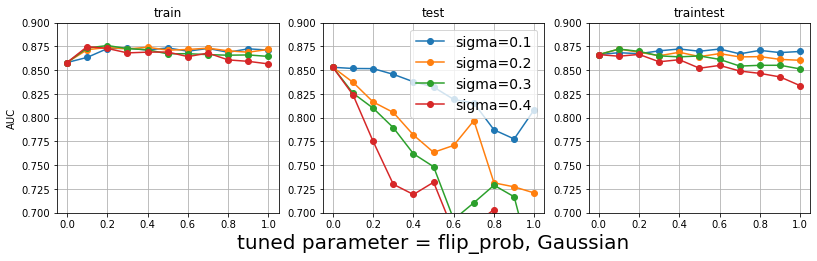}}
\caption{Fastai Gaussian scale and Bernoulli ratio}
\label{fastai_scale_tuning}
\end{center}
\vskip -0.2in
\end{figure}

\begin{figure}[ht]
\vskip 0.2in
\begin{center}
\centerline{\includegraphics[width=\columnwidth]{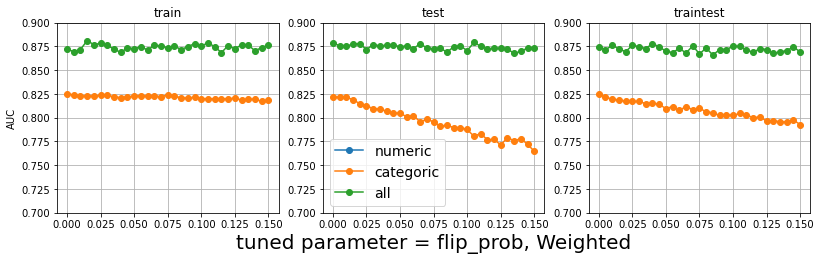}}

\centerline{\includegraphics[width=\columnwidth]{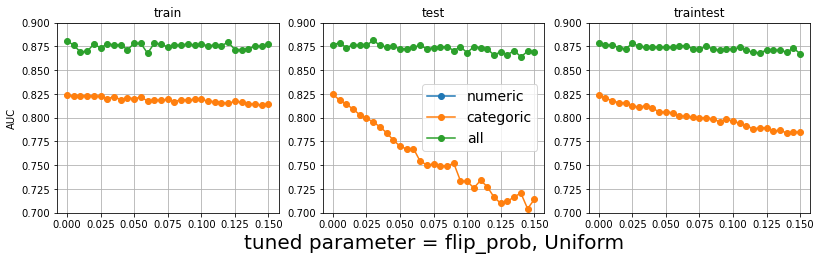}}
\caption{Fastai categoric injections}
\label{fastai_categoric_tuned_seperately}
\end{center}
\vskip -0.2in
\end{figure}

\begin{figure}[ht]
\vskip 0.2in
\begin{center}
\centerline{\includegraphics[width=\columnwidth]{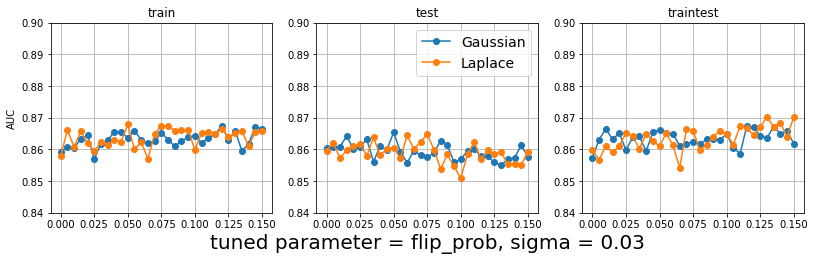}}
\centerline{\includegraphics[width=\columnwidth]{Exp04varynumericsigmawithlaplace_experimentpairs_0107_fastai_rescaled.png}}

\centerline{\includegraphics[width=\columnwidth]{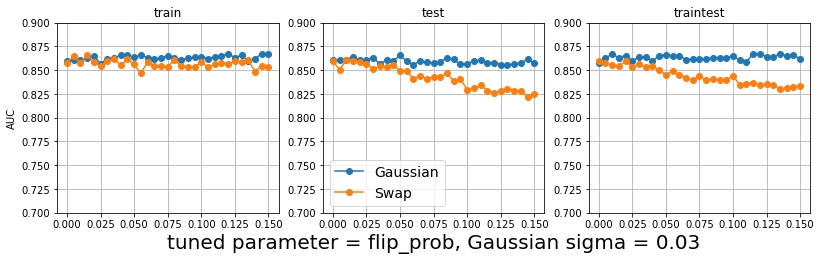}}

\centerline{\includegraphics[width=\columnwidth]{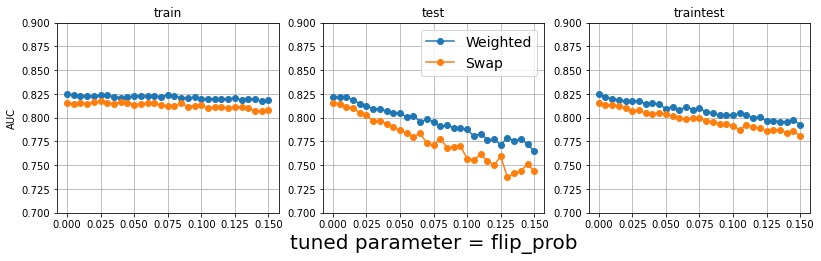}}

\centerline{\includegraphics[width=\columnwidth]{Exp06varycategoricflip_probuniform_experimentpairs_0107_fastai_rescaled.png}}


\caption{Fastai distribution comparisons}
\label{fastai_distribution_comparisons}
\end{center}
\vskip -0.2in
\end{figure}

The neural network scenario benefited from injecting noise to test data in conjunction with train data. The regularizing effect of the noise appeared to improve with increasing scale at low injection ratio [Fig \ref{fastai_Gaussian_tuned_seperately}]. When we increased the injection ratio (`flip\_prob') this tolerance of higher distribution scales (`sigma') faded [Fig \ref{fastai_scale_tuning}]. Unlike the numeric distribution sampling, categoric activation flips demonstrated an apparent linear performance degradation profile with increasing noise scale [Fig \ref{fastai_categoric_tuned_seperately}]. Note that this data set had many more numeric than categoric features, which we suspect made the categoric benchmarks appear disproportionately effected. Laplace distributed noise performed almost as well as Gaussian, it was very close [Fig \ref{fastai_distribution_comparisons}]. Note that Laplace's thicker tails means non-deterministic inference may be exposed to a broader range of scenarios.




\hyperref[A]{Table of Contents}

\newpage
\onecolumn
\twocolumn
\section{Sensitivity analysis - Catboost}
\label{J}

\textbf{Highlights:}

\begin{figure}[ht]
\vskip 0.2in
\begin{center}
\centerline{\includegraphics[width=\columnwidth]{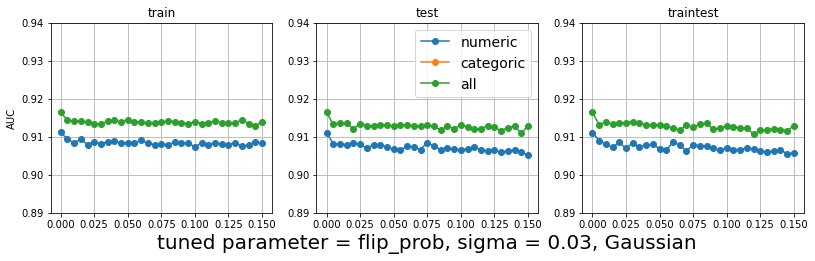}}

\centerline{\includegraphics[width=\columnwidth]{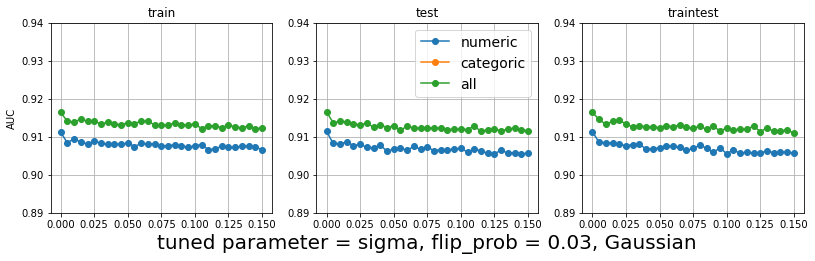}}
\caption{Catboost Gaussian injections}
\label{Catboost_Gaussian_tuned_seperately}
\end{center}
\vskip -0.2in
\end{figure}

\begin{figure}[ht]
\vskip 0.2in
\begin{center}
\centerline{\includegraphics[width=\columnwidth]{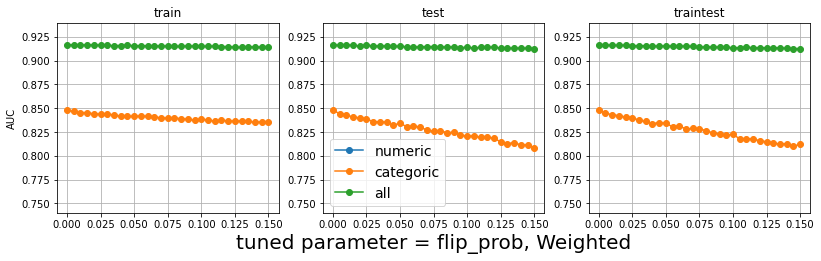}}

\centerline{\includegraphics[width=\columnwidth]{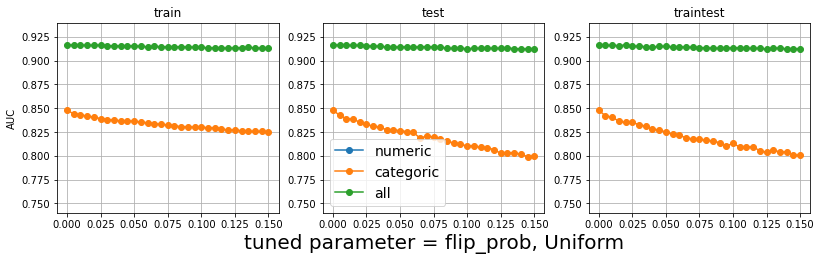}}
\caption{Catboost categoric injections}
\label{Catboost_categoric_tuned_seperately}
\end{center}
\vskip -0.2in
\end{figure}

\begin{figure}[ht]
\vskip 0.2in
\begin{center}
\centerline{\includegraphics[width=\columnwidth]{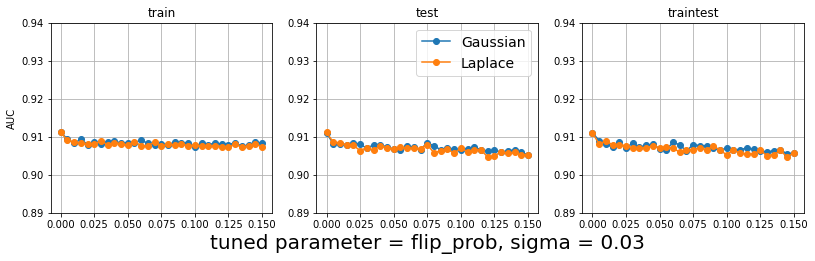}}
\centerline{\includegraphics[width=\columnwidth]{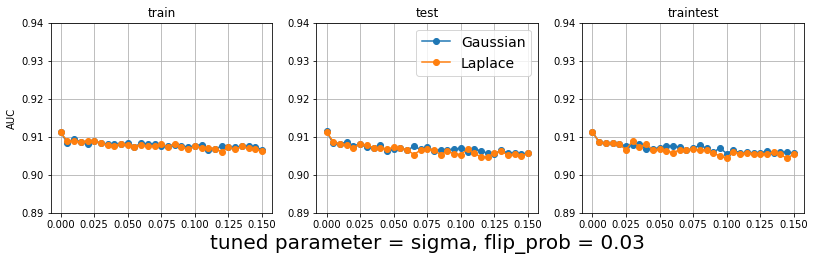}}

\centerline{\includegraphics[width=\columnwidth]{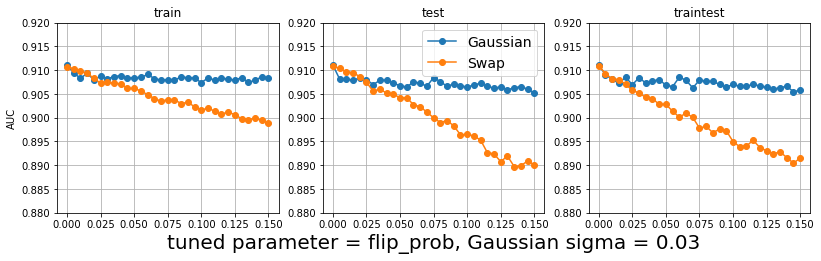}}

\centerline{\includegraphics[width=\columnwidth]{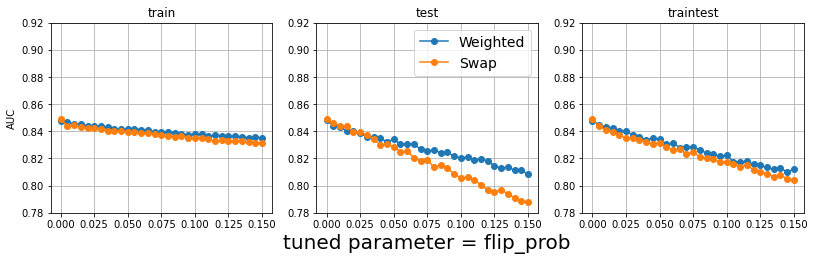}}

\centerline{\includegraphics[width=\columnwidth]{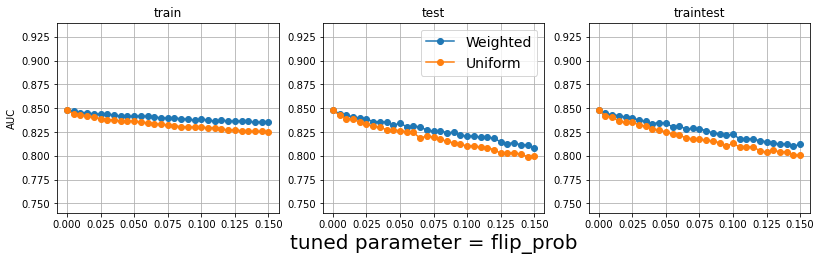}}


\caption{Catboost distribution comparisons}
\label{Catboost_distribution_comparisons}
\end{center}
\vskip -0.2in
\end{figure}

Catboost appeared to tolerate training data injections better than gradient boosting, and the traintest scenario for the most part closely resembled the test scenario. Unique to this library there was a characteristic drop at smallest numeric noise scale, and then further intensity was more benign [Fig \ref{Catboost_Gaussian_tuned_seperately}]. The categoric profile resembled a linear degradation with increasing ratio [Fig \ref{Catboost_categoric_tuned_seperately}]. Consistent with the other libraries, the categoric weighted sampling did not exactly line up with swap noise in test case [Fig \ref{Catboost_distribution_comparisons}], we expect this is because weighted sampling is fit to the training data while swap noise samples based on distribution of the target set, suggesting there was some amount of covariate shift in the validation set.

\hyperref[A]{Table of Contents}

\newpage
\onecolumn
\twocolumn
\section{Sensitivity analysis - XGBoost}
\label{K}

\textbf{Highlights:}

\begin{figure}[ht]
\vskip 0.2in
\begin{center}
\centerline{\includegraphics[width=\columnwidth]{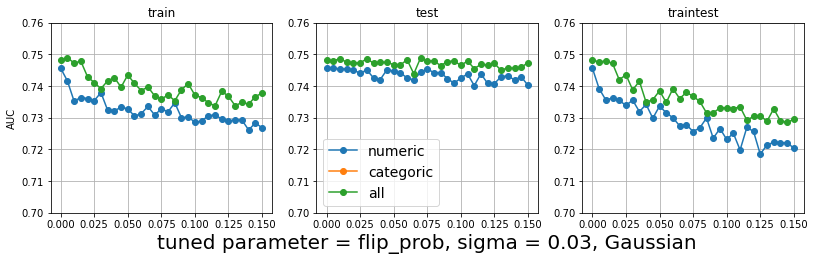}}

\centerline{\includegraphics[width=\columnwidth]{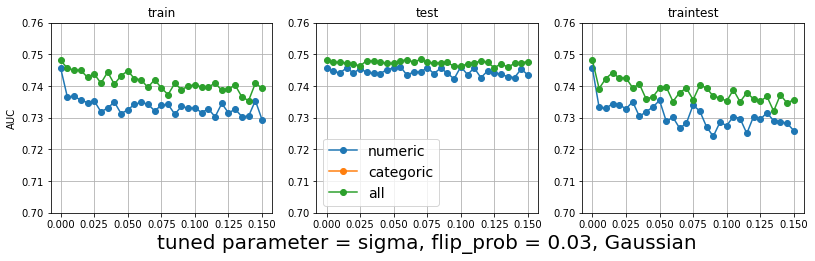}}
\caption{XGBoost Gaussian injections}
\label{XGBoost_Gaussian_tuned_seperately}
\end{center}
\vskip -0.2in
\end{figure}

\begin{figure}[ht]
\vskip 0.2in
\begin{center}

\centerline{\includegraphics[width=\columnwidth]{Exp040.4numericflip_prob_0107b_varytogether_tuned_sigma_xgboost_fliprobscenarios.png}}

\centerline{\includegraphics[width=\columnwidth]{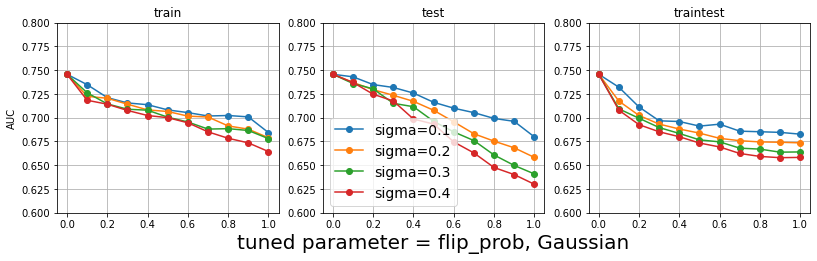}}
\caption{XGBoost Gaussian scale and Bernoulli ratio}
\label{XGBoost_scale_tuning}
\end{center}
\vskip -0.2in
\end{figure}

\begin{figure}[ht]
\vskip 0.2in
\begin{center}
\centerline{\includegraphics[width=\columnwidth]{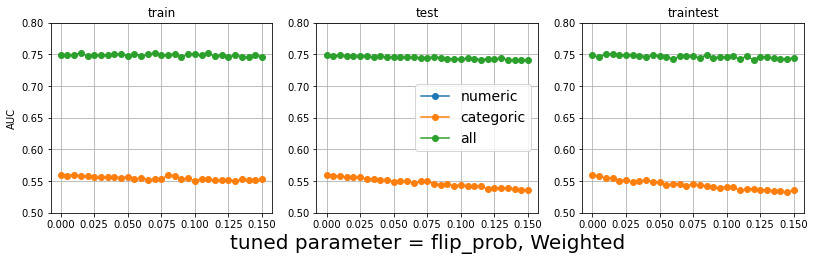}}

\centerline{\includegraphics[width=\columnwidth]{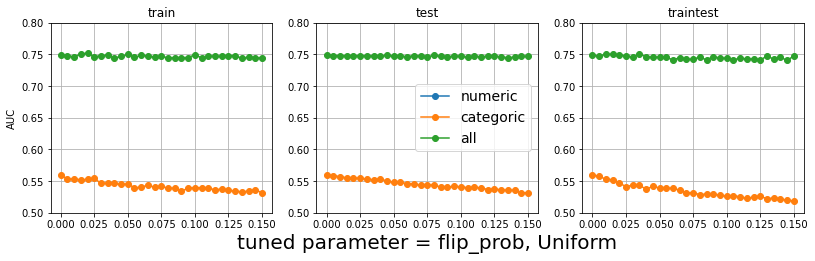}}
\caption{XGBoost categoric injections}
\label{XGBoost_categoric_tuned_seperately}
\end{center}
\vskip -0.2in
\end{figure}

\begin{figure}[ht]
\vskip 0.2in
\begin{center}
\centerline{\includegraphics[width=\columnwidth]{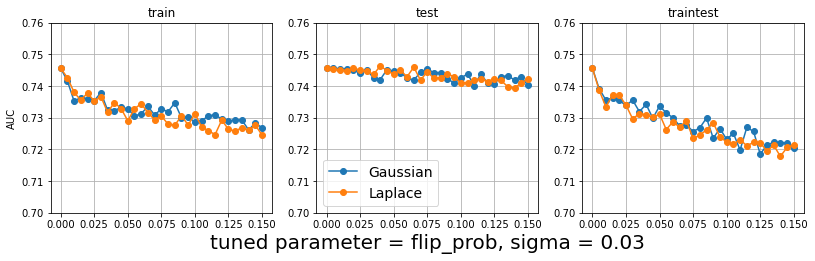}}
\centerline{\includegraphics[width=\columnwidth]{Exp04varynumericsigmawithlaplace_experimentpairs_0107_xgboost_rescaled.png}}

\centerline{\includegraphics[width=\columnwidth]{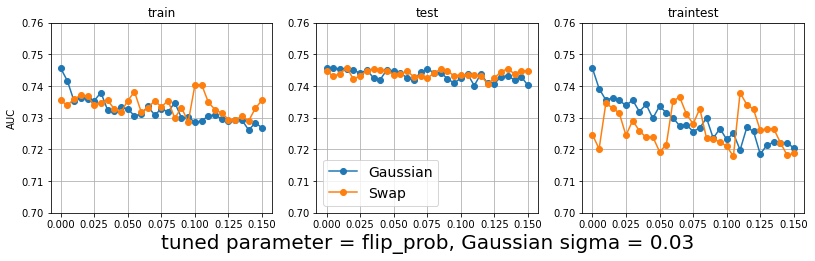}}


\centerline{\includegraphics[width=\columnwidth]{Exp06varycategoricflip_probuniform_experimentpairs_0107_xgboost_rescaled.png}}


\caption{XGBoost distribution comparisons}
\label{XGBoost_distribution_comparisons}
\end{center}
\vskip -0.2in
\end{figure}

The biggest takeaway here is that XGBoost does not like injections to train data, but injections only to test data for inference are tolerable with only small performance impact based on scale [Fig \ref{XGBoost_Gaussian_tuned_seperately}]. It appeared numeric features could tolerate increasing Gaussian scale with sufficiently low injection ratio [Fig \ref{XGBoost_scale_tuning}]. A subtle result, and don’t know if this would consistently duplicate, is that for categoric injections [Fig \ref{XGBoost_categoric_tuned_seperately}], when injecting to just test data, Weighted outperformed Uniform when looking at only categoric features, however when looking at all features (with the surrounding numeric features included), uniform slightly outperformed weighted. Numeric swap noise had a much closer performance to distribution sampling [Fig \ref{XGBoost_distribution_comparisons}] in comparison to neural networks [Fig \ref{fastai_distribution_comparisons}].

\hyperref[A]{Table of Contents}

\newpage
\onecolumn
\section{Intellectual property disclaimer}
\label{L}

Automunge is released under the MIT License. Full license details
available with code repository. Contact available via \url{https://www.automunge.com}. Copyright © 2022 — All Rights Reserved. Patent Pending.

\hyperref[A]{Table of Contents}

\end{document}